\documentclass{article}

\usepackage[final]{cpal_2025}

\usepackage{amsmath,amssymb,amsthm}
\usepackage{wasysym}
\usepackage{adjustbox}
\usepackage{xurl}
\usepackage{graphicx}
\usepackage{float}
\usepackage{booktabs}
\usepackage{algorithm}
\usepackage{algpseudocode}
\usepackage[most]{tcolorbox}
\usepackage{xcolor}
\usepackage{longtable}
\usepackage{listings}
\usepackage{microtype}    
\usepackage{inconsolata}  
\usepackage{wrapfig}
\usepackage{subcaption}

\usepackage{float}
\floatstyle{ruled}
\restylefloat{algorithm}
\usepackage{hyperref}

\definecolor{bioBack}{HTML}{cef5e8}
\definecolor{bioFrame}{HTML}{95f0d1}
\definecolor{histBack}{HTML}{bed5fa}
\definecolor{histFrame}{HTML}{74a3ed}
\definecolor{mathBack}{HTML}{e6acb9}
\definecolor{mathFrame}{HTML}{e3496a}

\tcbset{
  width=\linewidth,
  left=2pt,right=2pt,top=2pt,bottom=2pt,
  boxrule=0.4pt,
  coltitle=black,
  fonttitle=\bfseries,
}

\definecolor{dkgreen}{rgb}{0,0.6,0}
\definecolor{gray}{rgb}{0.5,0.5,0.5}
\definecolor{mauve}{rgb}{0.58,0,0.82}
\DeclareRobustCommand{\mcite}[1]{\textsuperscript{\tiny\citep{#1}}}

\makeatletter
\@ifundefined{citep}{\newcommand{\citep}[1]{\cite{#1}}}{}
\makeatother

\title{KNIGHT: Knowledge Graph-Driven Multiple-Choice Question Generation with Adaptive Hardness Calibration}

\author{%
  Mohammad Amanlou$^{1}$,
  Erfan Shafiee Moghaddam$^{2}$,
  Yasaman Amou Jafari$^{1,*}$,
  Mahdi Noori$^{1,*}$,
  Farhan Farsi$^{3}$,
  Behnam Bahrak$^{4}$
  \\
  $^{1}$University of Tehran \quad
  $^{2}$Independent Researcher \quad
  $^{3}$Amirkabir University of Technology \quad
  $^{4}$TEIAS Institute
  \\
  \texttt{Mohammad.amanlou@ut.ac.ir} \quad
  \texttt{erfanshm12@gmail.com} \quad
  \texttt{yasaman.jafary.a@ut.ac.ir}
  \\
  \texttt{mahdi.noori@ut.ac.ir} \quad
  \texttt{Farhan1379@aut.ac.ir} \quad
  \texttt{bahrak@teias.institute}
  \\
  $^{*}$Equal contribution.
}

\begin{document}

\maketitle

\begin{abstract}
With the rise of large language models (LLMs), they have become instrumental in applications such as Retrieval-Augmented Generation (RAG). Yet evaluating these systems remains bottlenecked by the time and cost of building specialized assessment datasets.
We introduce KNIGHT, an LLM-based, knowledge-graph-driven framework for generating multiple-choice question (MCQ) datasets from external sources. KNIGHT constructs a topic-specific knowledge graph, a structured, parsimonious summary of entities and relations, that can be reused to generate instructor-controlled difficulty levels, including multi-hop questions, without repeatedly re-feeding the full source text. This KG acts as a compressed, reusable state, making question generation a cheap read over the graph. We instantiate KNIGHT on Wikipedia/Wikidata, while keeping the framework domain- and ontology-agnostic.
As a case study, KNIGHT produces six MCQ datasets in History, Biology, and Mathematics. We evaluate quality on five criteria: fluency, unambiguity (single correct answer), topic relevance, option uniqueness, and answerability given the provided sources (as a proxy for hallucination). Results show that KNIGHT enables token- and cost-efficient generation from a reusable KG representation, achieves high quality across these criteria, and yields model rankings aligned with MMLU-style benchmarks, while supporting topic-specific and difficulty-controlled evaluation.

\end{abstract}

\section{Introduction}
Recent work identifies two main levers for LLM progress: model size and data \cite{kaplan2020scaling}. Because scaling parameters increases financial and environmental costs (e.g., CO\textsubscript{2} emissions) \cite{lakim-etal-2022-holistic,goel2025position}, attention is shifting to dataset curation; yet expert-quality datasets are expensive and slow to build, and in applied settings such as RAG and task-specific fine-tuning, public evaluation datasets remain scarce due to proprietary data despite available toolkits \cite{Lopatenko2024CompendiumLLMEvaluation,ragas2024}. Prior dataset-generation efforts exist \cite{vachev2022leaf,raina2022multiple,biancini2024multiple}, but there is still no widely adopted open-source framework that is reproducible and easy to implement. Moreover, standard MCQ benchmarks such as MMLU \cite{hendrycks2020mmlu} are largely static, difficult to update, provide limited instructor-level control over per-topic difficulty, and do not expose multi-hop structure for curriculum customization. In contrast, \textsc{KNIGHT} enables low-cost generation of topic-specific MCQ sets with user-controlled difficulty and explicit multi-hop design, while yielding model rankings aligned with MMLU-style \cite{hendrycks2021measuringmassivemultitasklanguage} benchmarks.

We introduce \textbf{\underline{K}nowledge-graph-driven \underline{N}atural \underline{I}tem \underline{G}eneration with \underline{A}daptive \underline{H}ardness \underline{T}uning} (\textbf{\textsc{KNIGHT}}), a fully automated framework for synthesizing large-scale MCQ datasets from external document collections and ontologies with controllable difficulty. Given a user topic $\tau$ (and optional prompt), \textsc{KNIGHT} runs four stages: (i) \emph{construct} a topic-specific knowledge graph (KG) via retrieval-augmented extraction \citep{lairgi2024itext2kg,lewis2020retrieval,guu2020realm}, where the KG is a compact, parsimonious summary of entities and relations distilled from the sources; (ii) \emph{generate} source-grounded MCQs by traversing multi-hop KG paths with configurable depth; (iii) \emph{calibrate} difficulty based on path length and abstraction, validated via entropy-based uncertainty measures and human error patterns; and (iv) \emph{filter} items with an LLM- and rule-based validator enforcing five criteria: grammar, single-correct-answer unambiguity, option uniqueness, answerability from evidence, and topicality \citep{fabbri2021emnlp,rejeleene2024towards}. 

\textsc{KNIGHT} integrates RAG-based extraction, KG-guided multi-hop generation, and LLM-based validation into a reusable, modular pipeline that caches a compact topic KG for efficient dataset creation, supports forward/reverse question modes, and uses human and entropy-based checks to mitigate imperfect answerability/difficulty proxies.

 \begin{wrapfigure}[10]{r}{0.40\textwidth}
    \vspace{-30pt}
    \centering
    \includegraphics[width=0.8\linewidth]{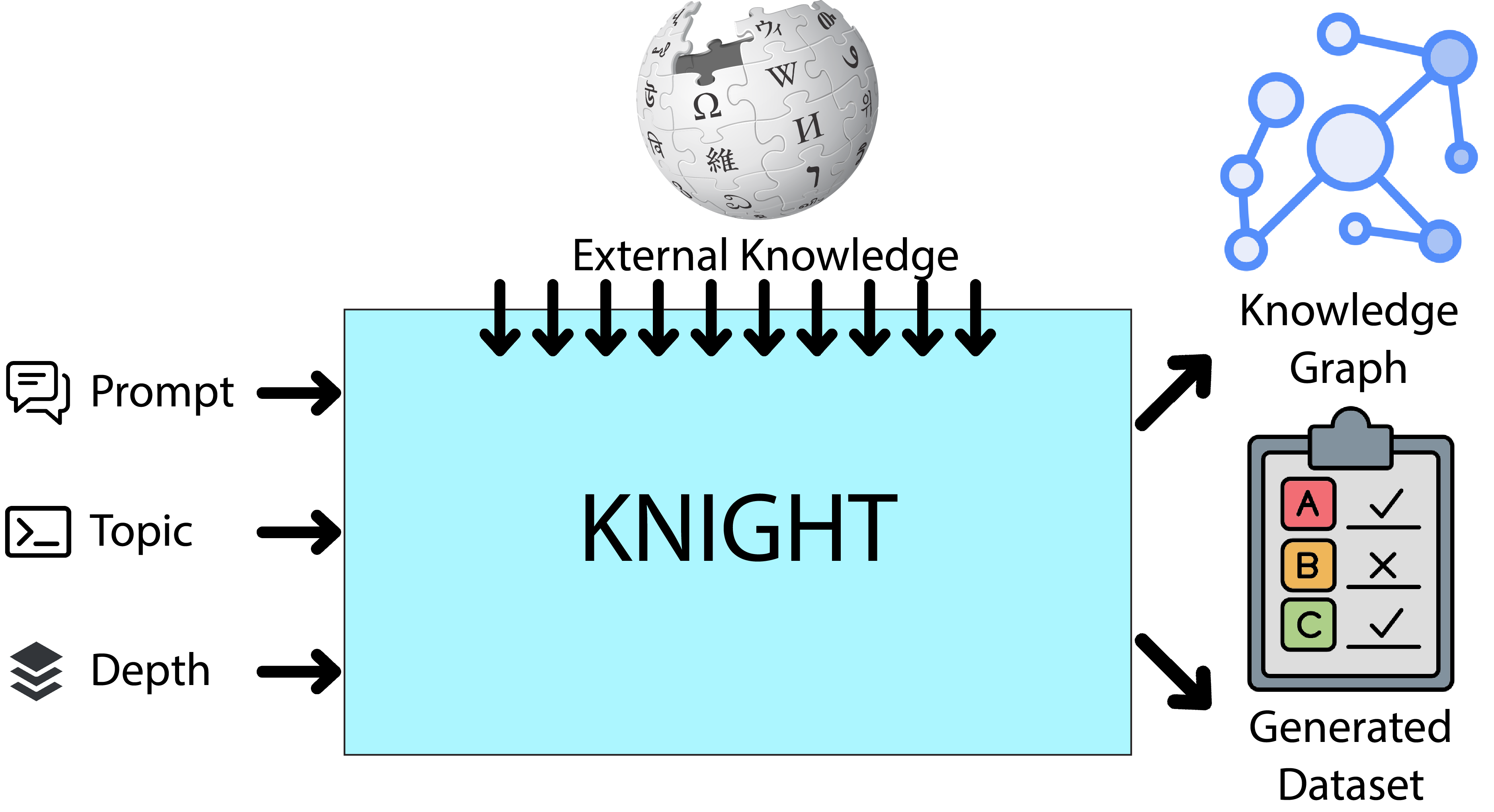}
    \caption{\textbf{\textsc{KNIGHT} High-level pipeline.} Given a prompt/topic and depth, KNIGHT retrieves evidence, builds a focused KG, generates MCQs, and filters them to produce the final dataset.}
    \vspace{-10pt}
    \label{fig:overview}
\end{wrapfigure}

We treat answerability from retrieved evidence as a proxy for generator hallucination: items judged unanswerable from the sources indicate unsupported or hallucinated content. Since the KG is built once per topic and reused across many generations, \textsc{KNIGHT} enables token-efficient, low-cost generation from a reusable KG representation versus naive prompting that repeatedly re-ingests long evidence contexts per question. We use \textbf{GPT-4o-mini} for all LLM calls throughout, yielding a cost- and token-aware evaluation setting.

We instantiate the type system $\phi$ on the Wikipedia/Wikidata ontology (though in principle $\phi$ can be defined over any domain ontology or enterprise schema) and introduce \textsc{KNIGHT}, a token-efficient, KG-driven pipeline for generating difficulty-controlled MCQ datasets from external sources and ontologies. As case studies of its flexibility and reusability (rather than standalone benchmark contributions), we construct six Wikipedia/Wikidata-based MCQ datasets spanning history, biology, and mathematics at two difficulty levels, and run \textbf{four ablations} alongside full \textsc{KNIGHT} using \textbf{five GPT-4o-mini configurations} (Plain, RAG, RAG+KG, RAG+Val, and full \textsc{KNIGHT}). This staged comparison isolates how grounding, KG guidance, and validation affect hallucination, distractor quality, and difficulty calibration via automatic, human, and entropy-based evaluations. Items are generated within minutes on Google Colab T4 and are grammatical and difficulty-calibrated (Section~\ref{sec:experiments}). \textsc{KNIGHT} reduces hallucinations relative to Plain and RAG (Section~\ref{sec:limitations}) and yields model rankings aligned with MMLU-style benchmarks, while being cheaper and easier to update than static MMLU-like test sets, supporting it as a scalable, low-cost benchmark generator.
Our code and package are publicly available on PyPI\footnote{\url{https://pypi.org/project/knight-mcq/}} and GitHub\footnote{\url{https://github.com/ErfanShm/knight-mcq}}.

\section{Related Work}

\paragraph{Knowledge Graph Construction.}
Constructing KGs from unstructured text typically uses multi-stage NLP pipelines (e.g. entity extraction/linking and relation extraction), often assuming a predefined schema and substantial supervision/training data \citep{zhong2023surveykgc,hogan2021knowledgegraphs}. Traditional systems commonly perform named entity recognition and model-based relation extraction to identify entities and relationships, but these often require predefined schemas and extensive training. Recent LLM-based methods reduce these requirements and improve portability: \citet{lairgi2024itext2kg} propose \textit{iText2KG}, a zero-shot incremental framework with LLM-powered entity/relation extraction for topic-independent KG construction; \citet{dessi2021generating} extract triples from scientific abstracts via NLP/text-mining and integrate them into a KG; and \citet{zhu2024llms} evaluate GPT-4 \cite{openai2023gpt4} on KG tasks, finding that it excels at reasoning, and introduce \textit{AutoKG}, a multi-agent LLM approach with external retrieval. Prompting has also improved relation extraction, e.g., Wikidata-informed prompts in \citet{layegh2024wiki}. While we instantiate our type system using Wikipedia/Wikidata as an ontology \citep{vrandecic2014wikidata}, the mapping function $\phi$ can in principle be defined over other domain ontologies or schemas (e.g., enterprise KGs or specialized KBs); here we evaluate only the Wikipedia/Wikidata instantiation and leave broader generalization to future work. Finally, in line with retrieval-augmented generation, RAG combines parametric LMs with retrieved knowledge bases; \citet{lewis2020retrieval} show it yields more specific, diverse, and factual outputs than parametric-only models.

\paragraph{Question Generation from Knowledge Graphs and Structured Data.}
Early work generates natural questions from KG triples using keyword extraction and RNNs \citep{reddy-etal-2017-generating}. Later methods go beyond single triples by encoding subgraphs with Graph2Seq and copy mechanisms \citep{chen2023toward}, and by using contextual KGs with answer-aware GATs for coherent multi-hop question generation \citep{li2023multi}. Difficulty control has also been studied explicitly: \citet{kumar-etal-2019-difficulty} condition multi-hop generation on estimated KG difficulty, while \citet{cheng-etal-2021-guiding} guide reasoning complexity via step-by-step rewriting. Beyond graph-centric pipelines, \textit{LIQUID} builds QA datasets directly from text through summarization, entity extraction, and question generation \citep{lee2023liquid}.

\paragraph{Evaluation and Filtering of Generated Questions.}
Ensuring the quality of generated questions requires multi-faceted evaluation, and recent work applies dedicated QA evaluation metrics. High-quality MCQs require multi-faceted evaluation: \citet{moore2024automatic} survey metrics including LM perplexity, lexical diversity, grammar error rates, cognitive complexity, and answerability to assess fluency, uniqueness, and inferability, while \citet{shypula2025evaluating} highlight semantic diversity gains from preference-tuned LLMs. Beyond these quality dimensions, \emph{factuality and safety} remain concerns: even strong LLMs can hallucinate and may exhibit biases, motivating automatic filtering and validation when generating educational content \citep{rejeleene2024}. Factuality is further challenged by the tendency of LLMs to answer confidently even when inputs are unanswerable, motivating explicit answerability checks as a practical proxy for hallucination control \citep{slobodkin-etal-2023-curious}. In RAG, recent work also evaluates whether systems correctly \emph{reject} unanswerable requests, complementing accuracy on answerable ones \citep{peng-etal-2025-unanswerability}. Finally, LLM-based review/validator pipelines can automatically assess MCQ validity across multiple criteria, reducing reliance on purely manual screening \citep{mucciaccia-etal-2025-automatic}.

Building on these lines, we propose \textsc{KNIGHT}, a unified \emph{end-to-end framework} that integrates KG construction, graph-driven question generation, and automatic quality filtering. A user-defined difficulty parameter controls graph depth to elicit multi-hop or higher-order items, while LLMs both generate and validate MCQs from KG paths. Compared to prior pipelines, our approach emphasizes reusable, token-efficient KG representations \emph{and} a comprehensive LLM-powered evaluation/validation stack, aiming to produce diverse, high-quality QA pairs with improved reliability for topic-specific question sets.

\section{System Design}\label{ssec:sysdes}
\subsection{Knowledge-Graph Constructor}
\label{ssec:kg-constructor}
\begin{figure*}[t]
    \centering
    \includegraphics[width=.95\textwidth]{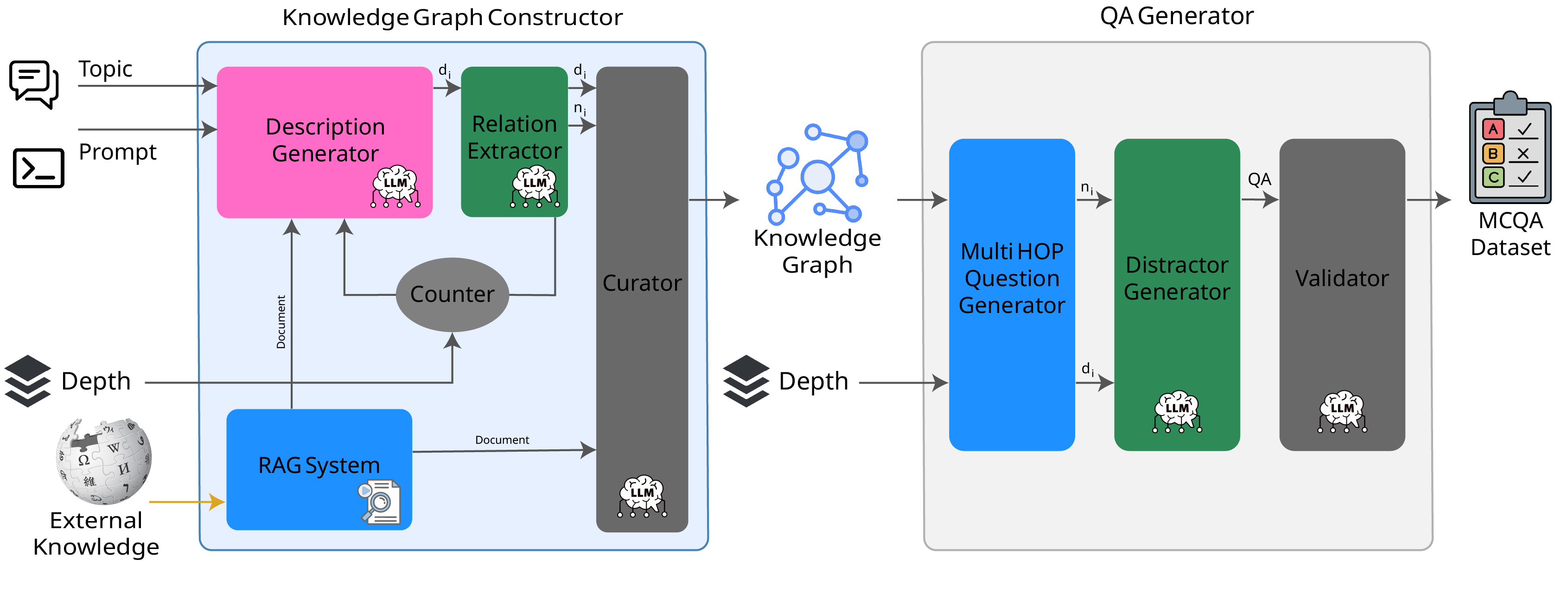}
    \caption{\textbf{KNIGHT architecture.} (Left) A topic/prompt-driven RAG pipeline retrieves evidence, extracts triples, and curates a compact KG under depth budget $d_{\max}$; (Right) multi-hop paths are sampled to generate questions/distractors and validated for evidence-grounded answerability to form the final MCQA dataset.}
    \label{fig:model_overview}
    \vspace{-22pt}
\end{figure*}

\noindent
\begin{minipage}[t]{0.56\linewidth}
\vspace{-0.8ex}
Given a user-specified topic $\tau$, optional prompt, and hardness budget $d_{\max}\in\mathbb{N}$,
the Knowledge-Graph Constructor builds a directed property graph $G=(V,E,\mathcal{R})$ with
canonicalized entities $v\in V$ and labeled edges $(v_h,r,v_t)\in E$, $r\in\mathcal{R}$.
It iterates a retrieve--generate--filter loop (Alg.~\ref{alg:graphgen}) combining external retrieval
with LLM reasoning; the backend is swappable (HuggingFace-compatible) via a config flag.

\vspace{0.25em}\noindent\textbf{Parsimonious, reusable representation.}
Once built, the KG can be cached and reused to generate many difficulty-controlled question sets
(varying hop length, formats, and targets) without re-feeding long source documents, amortizing the
one-time construction cost and improving token efficiency.
\paragraph{Evidence retrieval and description synthesis.}
We first retrieve a ranked context $\mathcal{D}=\{d_1,\dots,d_k\}$ from Wikipedia (or other open sources)
using dense passage retrieval and re-ranking~\citep{lewis2020retrieval,guu2020realm}.
Conditioned on $\tau$ (and 
\end{minipage}
\hfill
\begin{minipage}[t]{0.42\linewidth}
\vspace{-2.4ex}
\footnotesize
\captionsetup{type=algorithm,skip=2pt}

\hrule height 0.4pt
\vspace{0.25em}
\captionof{algorithm}{\textsc{GraphGenerator}---depth-bounded KG construction}
\label{alg:graphgen}
\vspace{0.15em}
\hrule height 0.4pt
\vspace{0.35em}

\begin{algorithmic}[1]
  \Require seed topic/entity $v_0$, depth limit $d_{\max}$
  \State $G\gets$ empty graph; $G.\text{addNode}(v_0)$
  \State $Q\gets[(v_0,0)]$ \Comment{FIFO over (node,depth)}
  \While{$Q\neq\emptyset$}
    \State $(v,\ell)\gets\textsc{PopFront}(Q)$
    \State $\mathcal{D}\gets\textsc{Retrieve}(v)$
    \State $\delta\gets\mathcal{L}_{\mathrm{desc}}(v,\mathcal{D})$
    \State $R\gets\mathcal{L}_{\mathrm{rel}}(\delta)$ 
    \State $C\gets\textsc{Curate}(R)$
    \ForAll{$(v,r,u)\in C$}
      \State $G.\text{addNode}(u)$; $G.\text{addEdge}(v,r,u)$
      \If{$\ell+1\le d_{\max}$} \State $\textsc{PushBack}(Q,(u,\ell+1))$ \EndIf
    \EndFor
  \EndWhile
  \State \Return $G$
\end{algorithmic}

\vspace{0.35em}
\hrule height 0.4pt
\end{minipage}
\vspace{-3pt}
the optional prompt) and $\mathcal{D}$, the Description Generator
$\mathcal{L}_{\mathrm{desc}}$ produces a structured eight-point gloss $\delta$ (Appendix~\ref{sec:prompts}).
\paragraph{Triple induction and deduplication.}
The Relation Extractor maps $\delta$ to a triple set $R(\delta)$ (Eq.~\ref{eq:triple-set}) and removes near-duplicates using a Levenshtein filter with threshold $\lambda_{\max}$~\citep{lairgi2024itext2kg,dessi2021generating}, then passes the remaining candidates to the Curator.
\paragraph{Curation, pruning, and depth control.}
The Curator applies (i) type checks (instantiated here with Wikidata), (ii) NLI-based consistency checks between node glosses and relation statements~\citep{nie2020adversarial, bowman2015large,williams2018broad}, and (iii) content-policy screening following prior safety analyses~\citep{rejeleene2024towards}. Graph expansion proceeds breadth-first and stops at depth $d_{\max}$, yielding the bounded neighborhood $V_{d_{\max}}=\{\,v \mid \mathrm{dist}_{G}(v_0,v)\le d_{\max}\,\}$ (Eq.~\ref{eq:depth-ball}), which matches the KG scope to the downstream MCQ generator (Section~\ref{ssec:mcq-generator}).

\paragraph{KG-1: Retrieval-Augmented Description Synthesis.}\label{sec:description-synthesis}
Figure~\ref{fig:model_overview} shows the first stage: producing an eight-point gloss $\delta(v_{0})$ for the seed $v_{0}$ via a \emph{rank-and-generate} RAG pipeline~\citep{lewis2020retrieval,guu2020realm} with (i) a dense retriever $\mathcal{R}_{\mathrm{enc}}$ (Contriever base; \citealp{Izacard2022}) encoding topic $\tau$ and scoring against a BM25-filtered corpus \citep{Robertson2009}, and (ii) a cross-encoder re-ranker $\mathcal{R}_{\mathrm{rer}}$ (MiniLM-L12; \citealp{Reimers2020}) refining the top--50 to $k{=}5$ passages $\mathcal{D}_0{=}\{d_{1},\dots,d_{5}\}$ with scores $s(d_i)\!\in\![0,1]$. Each retained passage $d_i$ is injected into system-prompted \textbf{GPT-4\textit{o}-mini}, yielding a candidate description $d_i^{\star}=\mathcal{L}_{\mathrm{desc}}(\tau,d_i)$; to combine evidence we model the generation probability as a RAG mixture~\citep{lewis2020retrieval}:
\begin{equation}
  P(d \mid \tau)=\sum_{z\in\mathcal{D}_0} P_{\theta}(d \mid \tau,z)\underbrace{\frac{\exp s(z)}{\sum_{z'} \exp s(z')}}_{P_{\text{ret}}(z \mid \tau)},
  \label{eq:rag}
\end{equation}
where $P_{\theta}$ is parameterised by GPT-4\textit{o}-mini; if $\mathcal{D}_0=\varnothing$ (no scores $>\!0.15$), we fall back to parametric generation $P(d\mid\tau){=}P_{\theta}(d\mid\tau)$. We retain a node gloss only if it is traceable to at least one retrieved passage, using:
\begin{equation}
  \gamma(\delta) =
  \begin{cases}
    1 & \exists z \in \mathcal{D}_0 : \texttt{overlap}(z, \delta) \ge \eta, \\[4pt]
    0 & \text{otherwise},  \quad \eta = 0.35,
  \end{cases}
  \label{eq:gamma}
\end{equation}
discarding $\gamma(\delta)=0$ descriptions; this makes persistent node content externally verifiable, mitigating hallucination risk \citep{ji2023hallucination}. The validated description $\delta(v_{0})$ is then forwarded to the Relation Extractor (\S\ref{sec:relation-extraction}), enabling breadth-first expansion up to depth~$d_{\max}$ (Algorithm~\ref{alg:graphgen}).

\paragraph{KG-2: Triple Induction via Relation Extraction.}\label{sec:relation-extraction}
Given a stored gloss $\delta$ in \textsf{description}, we distil explicit facts using the extractor $\mathcal{L}_{\text{rel}}$, implemented with \textbf{GPT-4\textit{o}-mini}\footnote{Checkpoint \texttt{gpt\_4o\_mini\_2024\_05}.}, which is prompted to emit a JSON list of $(\textit{head},\textit{relation},\textit{tail})$ triples. Formally, $R(\delta)=\{(h,r,t)\mid h,t\in\mathcal{E},\, r\in\mathcal{R}\}$,\label{eq:triple-set}
where $\mathcal{E}$ is the dynamic entity inventory and $\mathcal{R}$ a controlled relation schema
(cf.\,\citealp{lairgi2024itext2kg,dessi2021generating}). A Levenshtein filter with threshold $\lambda_{\max}$ removes near-duplicate triples before insertion.

\paragraph{KG-3: Depth-Controlled Expansion (token-efficient).}
Graph growth is bounded by the hardness budget $d_{\max}$: we run breadth-first expansion with a FIFO queue over $(v,\ell)$ (Algorithm~\ref{alg:graphgen}), re-applying KG-1--KG-2 while $\ell<d_{\max}$, and halt with the visited set $V_{d_{\max}}=\{v \mid \mathrm{dist}_{G}(v_0,v)\le d_{\max}\}$\label{eq:depth-ball}, ensuring no node exceeds the user-defined cognitive radius.

\paragraph{KG-4: Curation and pruning.}
Each candidate triple $(h,r,t)$ is filtered by $\phi$ using (i) ontology-based type agreement (here: Wikidata)~\citep{vrandevcic2014wikidata}, (ii) NLI entailment consistency between $\delta(h)$, $\delta(t)$ and the relation phrase~\citep{nie2020adversarial}, and (iii) content-policy compliance~\citep{rejeleene2024towards}; we retain the edge iff $\phi(h,r,t)=\textsc{True}$. $\phi$ can be instantiated with other domain ontologies/schemas beyond Wikipedia/Wikidata.

\subsection{MCQ Generator}\label{ssec:mcq-generator}
Given a validated KG $G$, we generate difficulty-calibrated multi-hop MCQs in two stages: \textbf{MCQ-1} (path-conditioned synthesis) and \textbf{MCQ-2} (validation/filtering). Both use same \textit{GPT-4\textsubscript{o-mini}}, while the decoder is swappable via configuration.

\paragraph{MCQ-1: Multi-Hop MCQ Synthesis.}
For each seed $v_0\in V$, we enumerate length-$d$ forward/reverse paths (each hop in $E$), e.g., $P: v_0\xrightarrow{r_1}v_1\cdots\xrightarrow{r_d}v_d$ (or the reverse orientation). We verbalize $P$ into a compact context template $T(P)$ by concatenating node glosses $\{\delta(v_i)\}_{i=0}^{d}$ and relation labels $\{r_i\}_{i=1}^{d}$, then prompt $\mathcal{L}_{\text{q}}$ to output an MCQ tuple $M_P=(q_P,a_P,D_P)$ with a single-sentence stem $q_P$, key $a_P$, and three semantically proximate distractors $D_P$~\citep{yu2024distractor,haladyna2002guidelines}; distractor quality is evaluated via entropy signals and human audits (\S\ref{appendix:expert_analysis}).

\paragraph{MCQ-2: MCQ Validation \& Filtering.}\label{sec:mcq-validation}
Each candidate $M_P$ is scored by a validator $\mathcal{L}_{\text{val}}$ on five criteria adapted from item-writing best practices~\citep{alfertshofer2024analyzing,xu2024can}: (i) grammatical fluency, (ii) single-key correctness, (iii) option uniqueness, (iv) answer derivability from the provided evidence (i.e., $T(P)$ and retrieved sources), and (v) topic relevance (when fixed). We retain an item iff all criteria pass,
$\text{keep}(M_P)=\big[\bigwedge_{k=1}^{5}\text{criterion}_k(M_P)=\textsc{True}\big]$,
discarding the rest. This LLM-as-critic loop improves factual fidelity and pedagogical validity of synthetic questions~\citep{alfertshofer2024analyzing,bean2023large,xu2024can}. Retained items are serialized as JSONL with provenance metadata $\langle v_0,d,P,\text{orientation}\rangle$.

\vspace{-5pt}

\section{Experiments}\label{sec:experiments}

\vspace{-5pt}
\subsection{Datasets}\label{subsec:datasets}
\vspace{-5pt}
We use six domain-specific multiple-choice (MCQ) datasets as case studies to evaluate KNIGHT across three subject areas (Biology, Mathematics, History) and two difficulty levels (Level~1, Level~3): \textit{Bio-1}, \textit{Bio-3}, \textit{Math-1}, \textit{Math-3}, \textit{Hist-1}, and \textit{Hist-3}.
The History datasets contain 241 MCQs at Level~1 and 697 at Level~3; the Biology datasets contain 323 MCQs at Level~1 and 970 at Level~3; and the Mathematics datasets contain 298 MCQs at Level~1 and 1063 at Level~3.
\vspace{-5pt}
\subsection{Experimental Setup and Baselines}\label{subsec:baseline}
\vspace{-5pt}
All systems use the same base generator, \textsc{GPT-4o-mini}. For fair, evidence-grounded comparison (and to reduce hallucination), all \emph{RAG-based} variants share the same Wikipedia retrieval step and use the retrieved passages as evidence context \citep{ji2023hallucination,lewis2020retrieval,asai2023selfrag}. We isolate component effects by toggling retrieval grounding (RAG), topic structure (KG), and post-hoc filtering (validator), yielding five configurations: \textbf{Plain} (no evidence), \textbf{RAG} (evidence only), \textbf{RAG+KG} (evidence + topic KG; no validator), \textbf{RAG+Val} (evidence + validator; no KG), and \textbf{\textsc{KNIGHT}} (evidence + KG-guided multi-hop structuring + validator + difficulty control; Sec.~\ref{ssec:sysdes}). For each topic--difficulty split, we generate $N{=}100$ MCQs per system with fixed decoding and aligned Level~1/Level~3 settings; KG-based variants additionally use the same constructed KG for comparability.

\subsection{What makes a ``good'' MCQ dataset?}\label{subsec:mcq-criteria}
\vspace{-5pt}
We evaluate five item-quality criteria: \textbf{linguistic quality} (well-formed, fluent text), \textbf{unambiguity} (exactly one correct key), \textbf{option uniqueness} (non-overlapping distractors), \textbf{answerability from source} (the key is derivable solely from the provided evidence), and \textbf{topic relevance} (semantic alignment with the declared topic).
In the following, we evaluate these criteria across all systems (Sec.~\ref{subsec:baseline}), and additionally report efficiency (generation speed) and difficulty calibration (Level~1 vs.\ Level~3).

\subsubsection{Linguistic Quality of Questions}
\vspace{-5pt}

\begin{table*}[t]
\centering
\resizebox{\textwidth}{!}{%
\begin{tabular}{lccccc ccccc ccccc}
\toprule
& \multicolumn{5}{c}{Grammar Accuracy $\uparrow$}
& \multicolumn{5}{c}{Fluency-automatic $\uparrow$}
& \multicolumn{5}{c}{Fluency-human $\uparrow$} \\
\cmidrule(lr){2-6}\cmidrule(lr){7-11}\cmidrule(lr){12-16}
Topic
& Plain & RAG & RAG+KG & RAG+Val & \textsc{KNIGHT}
& Plain & RAG & RAG+KG & RAG+Val & \textsc{KNIGHT}
& Plain & RAG & RAG+KG & RAG+Val & \textsc{KNIGHT} \\
\midrule
History & 0.9994 & 0.9993 & 0.9992 & 0.9990 & 0.9989  & 0.9498 & 0.9582 & 0.9536 & 0.9549 & 0.9581  & 4.8/5 & 4.7/5 & 4.7/5 & 4.7/5 & 4.8/5 \\
Biology & 0.9992 & 0.9994 & 0.9995 & 0.9989 & 0.9998  & 0.9702 & 0.9653 & 0.9681 & 0.9591 & 0.9626  & 4.9/5 & 4.9/5 & 4.8/5 & 4.8/5 & 4.9/5 \\
Math    & 0.9978 & 0.9986 & 0.9981 & 0.9983 & 0.9991  & 0.9711 & 0.9671 & 0.9658 & 0.9602 & 0.9685 & 4.9/5 & 4.8/5 & 4.7/5 & 4.8/5 & 4.7/5 \\
\bottomrule
\end{tabular}%
}
\caption{Linguistic quality aggregated over Levels. Systems: Plain GPT-4o-mini, GPT-4o-mini+RAG, GPT-4o-mini+RAG+KG, GPT-4o-mini+RAG+Validator, and \textsc{KNIGHT}.}
\label{tab:quality_metrics}
\end{table*}

We assess linguistic quality to avoid surface-form confounds along three axes: \emph{grammatical correctness}, \emph{fluency}, and \emph{question-length diversity}. \textbf{Grammar} is computed with LanguageTool \citep{languagetool,language_tool_python} as $\text{Grammar Quality}(q)=1-\frac{E}{W}$, where $W$ is the number of words and $E$ the detected errors. \textbf{Fluency} is measured both automatically with LangCheck \citep{langcheck}, whose scores correlate with human judgments \citep{fabbri2021emnlp}, and by CEFR C1/C2 annotators who rate $n{=}100$ randomly ordered items per dataset on a 5-point Likert scale (Appendix~\ref{appendix:expert_analysis}). \textbf{Length diversity} is analyzed via question-length distributions \citep{bean2023large,xu2024can} (Appendix~\ref{app:Question Length Diversity}). Table~\ref{tab:quality_metrics} reports grammar accuracy and fluency (automatic/human), aggregated over Levels, across topics and systems; overall linguistic quality is uniformly high, suggesting later differences primarily reflect grounding, structure, and validation rather than surface-form artifacts.
\vspace{-5pt}

\subsubsection{Unambiguity, Answerability, and Option Uniqueness}\label{subsec:validity-criteria}
\vspace{-5pt}
MCQ validity hinges on three properties: \textbf{unambiguity} (exactly one correct key), \textbf{evidence-grounded answerability} (the key is derivable from the provided evidence), and \textbf{non-overlapping distractors} (options are not near-duplicates). Violations inflate chance performance, undermine construct validity, and reduce score interpretability \citep{haladyna2002guidelines,downing2005validity,rodriguez2005three}.
\vspace{-5pt}

\paragraph{Human evaluation (Appendix~\ref{appendix:expert_analysis}).}
For each split (topic$\times$difficulty), blinded domain experts audited $n{=}100$ items per system and flagged four error types: \textsc{REPEATED}, \textsc{SINGLE\_KEY}, \textsc{OPTION\_UNIQUENESS}, and \textsc{ANSWERABLE} (key not justifiable from supplied evidence). We report counts per 100 items in Table~\ref{tab:answerability_analysis} (lower is better); evidence for judgments matches system inputs (none for \textbf{Plain}; retrieved passages for \textbf{RAG}/\textbf{RAG+Val}; passages+KG context for \textbf{RAG+KG}/\textsc{KNIGHT}).
\vspace{-5pt}

\paragraph{Answerability as a hallucination proxy.}
We treat \textsc{ANSWERABLE} violations as hallucination proxies: if the key is not derivable from the supplied evidence, the item is effectively ungrounded. Accordingly, \textbf{Plain} shows substantially higher \textsc{ANSWERABLE} counts, underscoring the role of retrieval grounding.
\paragraph{Results and component-wise patterns.}
Table~\ref{tab:answerability_analysis} shows that \textsc{KNIGHT} yields the cleanest item banks overall (low repetition, fewer ambiguity errors, stronger distractor separability, and fewer unanswerable items) across both Level~1 and Level~3; importantly, Level~3 does not substantially increase violations, suggesting difficulty control without sacrificing validity. full \textsc{KNIGHT} (KG structuring + validation + difficulty control) achieves the strongest validity profile under both difficulty settings.
\vspace{-5pt}

\begin{table*}[t]
\centering
\resizebox{\textwidth}{!}{%
\begin{tabular}{lccccc ccccc ccccc ccccc}
\toprule
& \multicolumn{5}{c}{\textsc{REPEATED} $\downarrow$}
& \multicolumn{5}{c}{\textsc{SINGLE\_KEY} $\downarrow$}
& \multicolumn{5}{c}{\textsc{OPTION\_UNIQUENESS} $\downarrow$}
& \multicolumn{5}{c}{\textsc{ANSWERABLE} $\downarrow$} \\
\cmidrule(lr){2-6}\cmidrule(lr){7-11}\cmidrule(lr){12-16}\cmidrule(lr){17-21}
Split
& Plain & RAG & RAG+KG & RAG+Val & \textsc{KNIGHT}
& Plain & RAG & RAG+KG & RAG+Val & \textsc{KNIGHT}
& Plain & RAG & RAG+KG & RAG+Val & \textsc{KNIGHT}
& Plain & RAG & RAG+KG & RAG+Val & \textsc{KNIGHT} \\
\midrule
History (L1) & 20 & 19 & 15 & 2  & 0  & 10 & 4 & 6 & 3  & 2  & 8 & 5 & 5 & 5  & 3  & 26 & 10 & 8 & 10 & 6 \\
Biology (L1) & 19 & 17 & 16 & 3  & 1  & 16 & 5 & 6 & 3  & 1  & 4 & 3 & 5 & 5 & 2  & 19 & 8  & 7 & 9  & 4 \\
Math (L1)    & 13 & 14 & 14 & 1  & 0  & 11 & 4 & 6 & 4  & 2  & 5 & 4 & 4 & 4  & 2  & 20 & 8  & 6 & 8  & 5 \\
History (L3) & 14 & 11 & 9 & 1  & 1  & 15 & 6 & 7 & 7  & 2  & 7 & 5 & 5 & 4  & 3 & 24 & 13 & 9 & 12 & 6 \\
Biology (L3) & 15 & 13 & 12 & 1  & 1  & 14 & 5 & 6 & 5  & 2  & 6 & 4 & 4 & 3  & 1  & 21 & 10 & 7 & 8  & 4 \\
Math (L3)    & 10 & 10 & 9 & 0  & 0  & 13 & 6 & 5 & 4  & 3  & 7 & 5 & 4 & 2  & 2  & 28 & 12 & 9 & 7  & 6 \\
\bottomrule
\end{tabular}%
}
\caption{Human audit flags per 100 items (lower is better). All systems use \textsc{GPT-4o-mini}: Plain, RAG, RAG+KG, RAG+Val, and \textsc{KNIGHT}.}
\label{tab:answerability_analysis}
\vspace{-5pt}
\end{table*}

\subsubsection{Topic Relevance}
\vspace{-5pt}
We assess topical alignment using two complementary signals: (i) zero-shot MNLI-style entailment \citep{williams2018broad}, treating the topic as premise and the question as hypothesis; and (ii) a large LLM in few-shot mode following standard NLG practice \citep{bowman2015large,williams2018broad} with prompting exemplars \citep{brown2020language}. For topic $T$ and question $q$,
\begin{equation}
S(q,T)=P(\text{entailment}\mid \text{premise}=T,\ \text{hypothesis}=q).
\label{eq:entailment_score}
\end{equation}
We additionally report expert \textsc{TOPIC} flags and an \emph{off-topic rate} computed from the union of the two automated checks (Table~\ref{tab:combined_topic_relevance}).

Table~\ref{tab:combined_topic_relevance} shows that \textsc{KNIGHT} maintains strong topical alignment across topics and difficulty levels: entailment and LLM-based relevance remain high, off-topic rates are low, and expert \textsc{TOPIC} flags are rare. Overall, these results indicate that the generated MCQs stay on-topic, enabling subsequent analyses to focus on validity, distractor quality, and difficulty calibration rather than topic drift.

\begin{table*}[t]
\centering
\resizebox{\textwidth}{!}{%
\begin{tabular}{lccccc ccccc ccccc ccccc}
\toprule
& \multicolumn{5}{c}{Topic Relevance Score (Entailment) $\uparrow$}
& \multicolumn{5}{c}{Topic Relevance (LLM) $\uparrow$}
& \multicolumn{5}{c}{Human TOPIC Flags $\downarrow$}
& \multicolumn{5}{c}{Off-topic Rate (LLM $\cap$ Entailment) $\downarrow$} \\
\cmidrule(lr){2-6}\cmidrule(lr){7-11}\cmidrule(lr){12-16}\cmidrule(lr){17-21}
Split
& Plain & RAG & RAG+KG & RAG+Val & \textsc{KNIGHT}
& Plain & RAG & RAG+KG & RAG+Val & \textsc{KNIGHT}
& Plain & RAG & RAG+KG & RAG+Val & \textsc{KNIGHT}
& Plain & RAG & RAG+KG & RAG+Val & \textsc{KNIGHT} \\
\midrule
History (L1) & 0.9432 & 0.9938 & 0.8080 & 0.9945 & 0.8214  & 0.8318 & 0.9116 & 0.7301 & 0.9218 & 0.7692  & 7 & 5 & 17 & 4 & 8  & 6\% & 4\%  & 19.3\% & 3\% & 10.6\% \\
Biology (L1) & 0.9156 & 0.9885 & 0.7962 & 0.9888 & 0.9053  & 0.8611 & 0.9542 & 0.7260 & 0.9619 & 0.7979  & 4 & 3 & 15 & 3 & 6  & 3\% & 1\%  & 18.0\% & 1\% & 5.5\% \\
Math (L1)    & 0.9219 & 0.9983 & 0.8103 & 0.9989 & 0.8852  & 0.8322 & 0.9092 & 0.8211 & 0.9514 & 0.8418  & 8 & 5 & 14 & 5 & 7  & 7\% & 4\%  & 14.7\% & 4\% & 7.3\% \\
History (L3) & 0.9086 & 0.9975 & 0.5765 & 0.9981 & 0.8974  & 0.8115 & 0.8884 & 0.5291 & 0.9384 & 0.8124  & 9 & 6 & 42 & 5 & 9  & 8\% & 6\%  & 34.4\% & 5\% & 8.1\% \\
Biology (L3) & 0.9309 & 0.9975 & 0.5971 & 0.9977 & 0.8832  & 0.8575 & 0.9205 & 0.5430 & 0.9381 & 0.8966  & 8 & 5 & 39 & 4 & 3  & 5\% & 3\%  & 22.4\% & 3\% & 3.4\% \\
Math (L3)    & 0.8996 & 0.9981 & 0.6203 & 0.9983 & 0.9849  & 0.8866 & 0.9307 & 0.5550 & 0.9438 & 0.8909  & 7 & 4 & 28 & 3 & 3  & 3\% & 2\%  & 19.0\% & 2\% & 2.2\% \\
\bottomrule
\end{tabular}%
}
\caption{Topic relevance (entailment and LLM; $\uparrow$), expert \textsc{TOPIC} flags ($\downarrow$), and off-topic rate as the intersection of automated checks (LLM $\cap$ Entailment; $\downarrow$) across systems (all with \textsc{GPT-4o-mini}).}
\label{tab:combined_topic_relevance}
\end{table*}
\vspace{-5pt}
\subsection{Generation Speed} \label{ssec:speed}
\vspace{-5pt}
We measure end-to-end wall-clock time per topic--difficulty split on commodity hardware (Google Colab: NVIDIA Tesla T4, 12 CPU cores; Appendix~\ref{app:generation-speed}). Level~1 completes in a few minutes (History: 212s, Math: 310s, Bio: 551s), while Level~3 remains practical (History: 852s, Math: 1226s, Bio: 2449s $\approx$ 41 min). These runtimes reflect \emph{dataset construction} (not a single exam) and enable fast, refreshable topic-specific MCQ banks compared to longer expert curation cycles for broad static benchmarks (e.g., MMLU-style suites).

KNIGHT is token- and cost-aware by design: the topic KG is built once per topic and cached, then reused to generate many variants (levels, hop lengths, and forward/reverse formats) without repeatedly re-feeding full source documents. Consequently, the \emph{end-to-end} token usage averages $\sim$600 \emph{total tokens per question} (prompt+completion across generation and validation stages) in our setup, whereas naive prompting and standard RAG-only baselines repeatedly inject longer evidence passages per item, inflating context length and latency. Caching amortizes the one-time KG construction cost and keeps the marginal cost of producing additional datasets low.
\vspace{-5pt}

\section{Discussion}
\subsection{Evaluation of Distractor Quality via Predictive Entropy}
\label{subsec:distractor-entropy}
High-quality four-option MCQs require distractors that \emph{compete} with the key without introducing ambiguity: weak distractors make items trivial, while misleading distractors can increase \textsc{SINGLE\_KEY} and \textsc{ANSWERABLE} violations in human audits (Sec.~\ref{subsec:validity-criteria}). Following \citet{kim2024clickbenchmarkdatasetcultural}, we quantify distractor ``pull'' via predictive entropy over answer choices using a small fixed probe model (\texttt{LLaMA~3.2-3B-Instruct}). Given probe logits $\mathbf{z}=(z_A,z_B,z_C,z_D)$, we compute $p_i=\exp(z_i)/\sum_{j=1}^4\exp(z_j)$ for $i\!\in\!\{A,B,C,D\}$ \citep{touvron2023llama} and entropy $H=-\sum_i p_i\log p_i$. Higher $H$ indicates distractors receive non-trivial probability mass (stronger competition) and should coincide with lower probe accuracy when items are genuinely harder.

\begin{table*}[t]
\centering
\resizebox{\textwidth}{!}{%
\begin{tabular}{l rrrrr rrrrr rrrrr}
\toprule
& \multicolumn{5}{c}{Mean Entropy $H$ $\uparrow$}
& \multicolumn{5}{c}{Std.\ Dev.\ of $H$ $\uparrow$}
& \multicolumn{5}{c}{Probe Acc.\ (\%) $\downarrow$} \\
\cmidrule(lr){2-6}\cmidrule(lr){7-11}\cmidrule(lr){12-16}
Split
& Plain & RAG & RAG+KG & RAG+Val & \textsc{KNIGHT}
& Plain & RAG & RAG+KG & RAG+Val & \textsc{KNIGHT}
& Plain & RAG & RAG+KG & RAG+Val & \textsc{KNIGHT} \\
\midrule
History (L1) & 0.0 & 0.0106 & 0.0122 & 0.0098 & 0.0134  & 0.0 & 0.0058 & 0.0692 & 0.0046 & 0.0855  & 100.00 & 99.00 & 88.00 & 98.00 & 86.83 \\
Biology (L1) & 0.0 & 0.0038 & 0.0096 & 0.0046 & 0.0189  & 0.0 & 0.0007 & 0.0897 & 0.0004 & 0.0803  & 100.00 & 98.00 & 89.00 & 99.00 & 86.21 \\
Math (L1)    & 0.0 & 0.0021 & 0.0256 & 0.0039 & 0.0231  & 0.0 & 0.0004 & 0.0991 & 0.0005 & 0.1084  & 100.00 & 100.00 & 84.00 & 98.00 & 84.29 \\
History (L3) & 0.0 & 0.0 & 0.0435 & 0.0011 & 0.0489  & 0.0 & 0.0 & 0.1846 & 0.0002 & 0.1703  & 100.00 & 99.00 & 69.00 & 99.00 & 66.29 \\
Biology (L3) & 0.0 & 0.0017 & 0.0191 & 0.0039 & 0.0278  & 0.0 & 0.0003 & 0.1156 & 0.0003 & 0.1144  & 100.00 & 99.00 & 71.00 & 99.00 & 66.48 \\
Math (L3)    & 0.0 & 0.0 & 0.0699 & 0.0021 & 0.0826  & 0.0 & 0.0 & 0.2006 & 0.0001 & 0.2288  & 100.00 & 100.00 & 80.00 & 100.00 & 79.02 \\
\bottomrule
\end{tabular}%
}
\caption{\textbf{Distractor competition via predictive entropy.} Mean and standard deviation of predictive entropy ($H$) (higher $\uparrow$ = more competitive distractors) and probe accuracy (lower $\downarrow$ = harder items) using \texttt{LLaMA~3.2-3B-Instruct} as a fixed probe.}
\label{tab:entropy_stats}
\end{table*}

\noindent\textbf{Findings (Table~\ref{tab:entropy_stats}).}
\emph{(1) Difficulty signal (Level~1 $\rightarrow$ Level~3).} Across domains, \textsc{KNIGHT} shows the expected pattern: entropy increases from Level~1 to Level~3 while probe accuracy decreases, and Std.\ Dev.\ rises at Level~3, indicating a broader, more realistic hardness spread rather than tightly clustered difficulty. The same directionality is visible for KG-guided prompting (\textsc{RAG+KG}), reinforcing that $H$ tracks intended difficulty.

\noindent\emph{(2) Component-wise contrast.} \textsc{Plain} is degenerate, with near-zero entropy and $100\%$ probe accuracy, consistent with non-competitive distractors. Among grounded baselines, \textsc{RAG} and \textsc{RAG+Val} remain near-zero in $H$ with near-ceiling probe accuracy across splits, suggesting retrieval alone and validator-only filtering (without KG guidance) does not reliably induce close distractors. In contrast, \textsc{RAG+KG} substantially increases $H$ and lowers probe accuracy, indicating that KG-conditioned, path-/fact-driven prompting is the main driver of semantically proximate distractors. Full \textsc{KNIGHT} (RAG+KG+Validator) achieves the strongest overall competition profile (higher $H$ with lower probe accuracy) while remaining well-formed under the same validity constraints used in human audits (Sec.~\ref{subsec:validity-criteria}).

\noindent\emph{(3) Domain patterns and spread.} Math at Level~3 exhibits the highest entropy (and correspondingly lower probe accuracy), consistent with especially close distractors; History and Biology show the same qualitative Level~1$\rightarrow$Level~3 shift. Across topics, the larger Level~3 Std.\ Dev.\ for \textsc{RAG+KG} and \textsc{KNIGHT} suggests that KG-driven prompting yields not only harder items on average but also greater within-split difficulty diversity, whereas non-KG baselines cluster near triviality.

\subsection{Are \textsc{KNIGHT} datasets reliable, hard, and usable as benchmarks?}
\label{subsec:benchmark-utility}
Table~\ref{tab:knight_full_models_by_dataset_en} reports a controlled evaluation of multiple LLMs on \textsc{KNIGHT} (three domains $\times$ two difficulty levels) alongside standard MCQ benchmarks (\textsc{MMLU}~\citep{hendrycks2020mmlu}, \textsc{ARC}~\citep{clark2018arc}, \textsc{CSQA}~\citep{talmor2019commonsenseqa}, \textsc{RACE}~\citep{lai2017race}, \textsc{MedMCQA}~\citep{pal2022medmcqa}, \textsc{OpenBookQA}~\citep{mihaylov2018openbookqa}). We follow official (or de facto) evaluation scripts and adopt the Open-LLM-Leaderboard protocol of \citet{myrzakhan2024openllm} to mitigate MCQ selection bias and ensure cross-model comparability; we report both a \emph{\textsc{KNIGHT} average} (over domains/levels) and a \emph{benchmark average} (over external suites).

\begin{table}[!t]
\centering
\setlength{\abovecaptionskip}{2pt}
\setlength{\belowcaptionskip}{-2pt}
\small
\setlength{\tabcolsep}{5.5pt}
\renewcommand{\arraystretch}{1.12}
\resizebox{\textwidth}{!}{%
\begin{tabular}{@{}l*{14}{r}@{}}
\toprule
& \multicolumn{2}{c}{History} & \multicolumn{2}{c}{Biology} & \multicolumn{2}{c}{Math} & \multicolumn{1}{c}{\textsc{KNIGHT} Avg.} & \multicolumn{1}{c}{MMLU} & \multicolumn{1}{c}{ARC} & \multicolumn{1}{c}{CSQA} & \multicolumn{1}{c}{RACE} & \multicolumn{1}{c}{MedMCQA} & \multicolumn{1}{c}{OBQA} & \multicolumn{1}{c}{Bench Avg.} \\
\cmidrule(lr){2-3}\cmidrule(lr){4-5}\cmidrule(lr){6-7}
Model & L1 & L3 & L1 & L3 & L1 & L3 & & & & & & & & \\
\midrule
GPT-4o\mcite{openaigpt4o}                & 92.95 & 86.39 & 95.98 & 87.11 & 95.30 & 86.41 & 90.52\textsuperscript{1st} & 79.09 & 86.31 & 70.28 & 67.87 & 57.85 & 67.21 & 71.45\textsuperscript{1st} \\
Mistral Large\mcite{mistral_large}        & 92.19 & 84.16 & 95.18 & 86.84 & 95.07 & 84.10 & 89.59\textsuperscript{2nd} & 68.76 & 72.32 & 55.35 & 70.17 & 43.44 & 58.66 & 61.45\textsuperscript{2nd} \\
Llama3-70B-Instruct\mcite{meta_llama3}    & 92.12 & 85.15 & 94.12 & 86.08 & 95.03 & 84.67 & 89.53\textsuperscript{3rd} & 59.67 & 67.09 & 55.49 & 58.21 & 41.67 & 40.94 & 53.85\textsuperscript{3rd} \\
Claude 3 Haiku\mcite{anthropic_claude3}   & 91.95 & 81.64 & 95.05 & 83.40 & 93.97 & 81.60 & 87.70 & 57.35 & 63.89 & 55.87 & 57.20 & 40.57 & 42.32 & 52.87 \\
Qwen1.5 (1.8B)\mcite{qwen15}              & 76.76 & 72.45 & 79.88 & 72.89 & 75.84 & 71.43 & 74.88 & 9.99  & 15.84 & 31.13 & 34.91 & 4.70  & 20.37 & 19.49 \\
Gemma (2B)\mcite{gemma2b}                 & 38.59 & 30.73 & 45.51 & 31.27 & 43.62 & 40.45 & 38.36 & 17.52 & 23.93 & 27.40 & 14.32 & 4.57  & 14.26 & 17.00 \\
\addlinespace
Human (n=200)                             & 98.60 & 89.20 & 97.40 & 91.60 & 97.40 & 89.40 & 93.92 & --    & --    & --    & --    & --    & --    & --    \\
\bottomrule
\end{tabular}%
}
\caption{\textbf{Benchmark utility of \textsc{KNIGHT}.} Accuracy (\%) of multiple models on \textsc{KNIGHT}, alongside standard MCQ benchmarks. \textsc{KNIGHT Avg.} averages over domains and levels; \textsc{Bench Avg.} averages over external suites. Superscripts denote rank by the corresponding average.}
\label{tab:knight_full_models_by_dataset_en}
\end{table}
\vspace{-5pt}
\paragraph{Difficulty calibration and reliability.}
Across models (from GPT-4o to 2B-scale baselines), Level~3 accuracy is consistently lower than Level~1 within each domain, indicating a stable, model-agnostic difficulty separation. A 200-item human study mirrors this pattern (lower L3 vs.\ L1 while remaining high), supporting that items are well-posed (unambiguous keys, reasonable distractors) and that Level~3 is genuinely more demanding rather than error-prone.
\vspace{-5pt}
\paragraph{Convergent validity with established benchmarks.}
Beyond within-domain calibration, ranking under the \emph{\textsc{KNIGHT} average} closely matches the \emph{benchmark average} (GPT-4o highest, followed by Mistral Large and Llama3-70B-Instruct, with smaller models trailing), suggesting that \textsc{KNIGHT} captures difficulty factors predictive of general QA competence rather than overfitting to a narrow topic distribution or prompting style.
\vspace{-5pt}
\paragraph{Parsimony and cost--quality trade-off.}
\textsc{KNIGHT} is token- and cost-aware: the topic KG is constructed once and reused as a compact representation to generate many variants (different levels and item patterns) without repeatedly injecting long evidence passages. This yields low marginal cost and enables frequent benchmark refresh, whereas broad static suites require substantial expert effort and longer build cycles, and naive prompting pipelines re-consume long contexts each generation pass, increasing token and runtime overhead.
\vspace{-5pt}
\paragraph{\textsc{KNIGHT} vs.\ broad static suites.}
\textsc{KNIGHT} complements wide-coverage benchmarks such as \textsc{MMLU}: MMLU provides standardized breadth, whereas \textsc{KNIGHT} offers refreshable, topic-scoped evaluation with fine-grained difficulty control and explicit multi-hop structure, while preserving rank-order agreement with established suites.
\subsection{Ablation Study: Component-wise Impact}
\label{subsec:ablation}
We isolate the contribution of retrieval grounding, KG guidance, and validation using the staged systems in Sec.~\ref{subsec:baseline}
(\textsc{Plain}, \textsc{RAG}, \textsc{RAG+KG}, \textsc{RAG+Val}, and full \textsc{KNIGHT}).
We treat \textsc{ANSWERABLE} violations as a hallucination proxy (unsupported content under the system-provided evidence) and use predictive entropy (Table~\ref{tab:entropy_stats}) as an automatic signal of distractor competition and controlled hardness.
\paragraph{Retrieval grounding (Plain $\rightarrow$ RAG).}
Table~\ref{tab:answerability_analysis} shows that retrieval substantially reduces \textsc{ANSWERABLE} violations and stabilizes topical alignment (Table~\ref{tab:combined_topic_relevance}).
However, \textsc{RAG} still exhibits near-zero entropy with near-ceiling probe accuracy (Table~\ref{tab:entropy_stats}), suggesting that grounding alone does not reliably yield competitive distractors or meaningful difficulty separation.
\paragraph{Validation without KG (RAG $\rightarrow$ RAG+Val).}
Adding the validator mainly improves item validity: it sharply reduces \textsc{REPEATED}, \textsc{SINGLE\_KEY}, and \textsc{OPTION\_UNIQUENESS} errors and further lowers \textsc{ANSWERABLE} violations (Table~\ref{tab:answerability_analysis}), while preserving topicality (Table~\ref{tab:combined_topic_relevance}).
Yet, as in \textsc{RAG}, entropy remains near zero with near-ceiling probe accuracy (Table~\ref{tab:entropy_stats}), indicating that validation alone does not strengthen distractor competition or enforce calibrated hardness.
\paragraph{KG guidance without validation (RAG $\rightarrow$ RAG+KG).}
Conditioning generation on KG paths/facts is the main driver of stronger distractor competition: entropy rises markedly (especially at Level~3) and probe accuracy drops (Table~\ref{tab:entropy_stats}), consistent with semantically closer distractors.
Without filtering, \textsc{RAG+KG} can also increase topical drift and validity errors (Table~\ref{tab:combined_topic_relevance}; Table~\ref{tab:answerability_analysis}), motivating post-hoc constraints in the full pipeline.
\paragraph{Combined effect (RAG+KG $\rightarrow$ KNIGHT; RAG+Val $\rightarrow$ KNIGHT).}
Combining KG-guided structuring with validation yields the strongest overall validity profile: \textsc{KNIGHT} achieves the lowest violation rates, with consistently fewer unanswerable items across Level~1 and Level~3 (Table~\ref{tab:answerability_analysis}); importantly, moving to Level~3 does not substantially increase violations, suggesting difficulty control without degrading validity.
\textsc{KNIGHT} also maintains topicality (Table~\ref{tab:combined_topic_relevance}).
Finally, it preserves the KG-induced difficulty signal—higher entropy, lower probe accuracy, and increased dispersion at Level~3—while avoiding the unfiltered \textsc{RAG+KG} error patterns (Table~\ref{tab:entropy_stats}).
The contrast between \textsc{RAG+Val} and \textsc{KNIGHT} isolates the added value of KG structure \emph{given} the same validator: KG prompting provides an interpretable, reusable scaffold that induces competitive distractors and controllable hardness, while the validator enforces item-writing constraints and further reduces hallucination as measured by answerability.

\section{Conclusion}
We presented \textsc{KNIGHT}, a knowledge-graph--driven framework for \emph{token-efficient} and \emph{low-cost} generation of topic-scoped four-option MCQ datasets with \emph{controllable difficulty}. Across three domains and two difficulty settings, \textsc{KNIGHT} produces linguistically polished items while substantially improving validity over retrieval-only prompting: expert audits show fewer duplicates, fewer ambiguous (multi-key) questions, stronger option uniqueness, and markedly lower unanswerable items, treating \textsc{ANSWERABLE} violations as a proxy for hallucination. Predictive entropy further provides a practical, model-agnostic signal of distractor competitiveness and difficulty, aligning with both human and model performance.

Beyond the specific case studies built from Wikipedia/Wikidata, the main contribution is the framework itself: a reusable KG representation that can be constructed once per topic and then leveraged to generate many question variants (levels, hop patterns, formats) at low marginal cost. This makes \textsc{KNIGHT} complementary to broad static benchmarks: while those suites offer standardized coverage, \textsc{KNIGHT} enables rapid, refreshable, syllabus-aligned evaluation with explicit multi-hop structure and instructor-controlled difficulty.

Future work includes extending \textsc{KNIGHT} beyond single-answer MCQs, incorporating adaptive difficulty tuning via model feedback, and strengthening robustness through adversarial evaluation and explainability. We also plan to instantiate the framework on alternative ontologies and domains, and to explore cross-lingual and multimodal settings.

\bibliography{reference}

\appendix

\section{Limitations}\label{sec:limitations}
\paragraph{Model choice.}
\textsc{KNIGHT} is modular and can use different LLMs for different stages. In this paper, for cost and reproducibility, we use a single model (\textbf{GPT-4o-mini}) for all LLM calls; we do not perform an exhaustive, task-wise model selection study.

\paragraph{Data domain.} We selected History, Biology, and Mathematics to represent a broad spectrum of relational diversity, ranging from narrative-heavy event chains to abstract logical structures. However, our findings may not fully generalize to domains with low relational density, such as Physics~\cite{petersen2025electrokinetic} or Numerical computation~\cite{tarzjani2025computing}, where knowledge is often encoded in numerical constants and first-principle equations rather than explicit entity-relation triples. In such "calculation-heavy" domains, the graph-based grounding used by \textsc{KNIGHT} may provide less utility than in the highly interconnected domains studied here.

\paragraph{Residual hallucination.}
Grounding and validation substantially reduce unsupported content, but do not eliminate it. We operationalize hallucination via the \textsc{ANSWERABLE} audit flag (Sec.~\ref{subsec:validity-criteria}); while \textsc{KNIGHT} lowers this rate compared to baselines, it remains non-zero (Table~\ref{tab:answerability_analysis}).

\paragraph{Difficulty is multi-factorial.}
Our primary hardness control relies on KG-based signals (e.g., multi-hop structure / graph distance), which correlate well with observed difficulty, but linguistic complexity and domain prerequisites can also affect hardness.

\paragraph{Evaluation scope.}
We evaluate only the Wikipedia/Wikidata instantiation and three domains as case studies; extending the evaluation to other corpora and ontologies is left for future work.

\section{Ethics Statement}
\textsc{KNIGHT} is released to support research and educational use via transparent, reproducible, low-cost generation of topic-scoped MCQ datasets with controllable difficulty. As with any open-source content-generation tool, it may be misused (e.g., to produce misleading content); we therefore encourage responsible use consistent with research integrity and institutional guidelines.

Our study uses only publicly available sources (Wikipedia/Wikidata) and does not involve personal or sensitive data. Users applying \textsc{KNIGHT} to private corpora should ensure compliance with privacy and licensing requirements and avoid including personally identifiable information.

Expert annotation targeted item quality and contained no sensitive content; exemption from IRB review was determined according to institutional guidelines. Any AI assistant usage was limited to editorial and stylistic revisions and did not contribute to research design, data collection, or analysis.

\section{System Usage}
\label{sec:system-usage}

In this section we first outline installation, then present the two user interfaces
(API \& CLI), and finally detail KG construction, generation, and
validation.

\subsection{Installation and Configuration}
\label{ssec:install}

The framework targets \texttt{Python\,$\ge$\,3.11} and installs in a single step:
\begin{tcolorbox}[
  colback=gray!5,
  colframe=gray!40,
  boxrule=0.35pt,
  left=2pt,right=2pt,top=2pt,bottom=2pt]
\footnotesize
\texttt{\$\,pip install knight-framework}
\end{tcolorbox}

All transitively required libraries are \emph{version-pinned} in
\texttt{uv.lock}, notably
LangChain \cite{hwchase2025langchain},
spaCy~3 \citep{spacy2},
Transformers~4 \citep{wolf-etal-2020-transformers},
and the Neo4j \cite{neo4jDocs} Python
driver; this guarantees byte-identical
reproduction.

\paragraph{External services.}
A Neo4j~5.x instance provides persistent KG storage, accessed via the
Bolt protocol:
\begin{tcolorbox}[colback=gray!5,colframe=gray!40,boxrule=0.35pt]
\footnotesize
\texttt{\$\,export NEO4J\_URI=bolt://localhost:7687}\\
\texttt{\$\,export NEO4J\_USER=neo4j}\\
\texttt{\$\,export NEO4J\_PASS=\textit{<pwd>}}
\end{tcolorbox}
Memory can be used instead of Neo4j instance by passing \texttt{backend=\allowbreak"memory"} to the constructor which results in a non-persistent KG storage.

For item synthesis we default to GPT-4o-mini. The API key of LLM must be provided:
\begin{tcolorbox}[colback=gray!5,colframe=gray!40,boxrule=0.35pt]
\footnotesize
\texttt{\$\,export OPENAI\_API\_KEY=\textit{sk-••••}}
\end{tcolorbox}
Any HuggingFace-compatible decoder (e.g.,\ Llama-2
\citep{touvron2023llama}) can be hot-swapped by setting
\texttt{lm\_backend="hf"}.

\subsection{Unified Workflow (API \& CLI)}
\label{ssec:interfaces}

Both interfaces expose identical functionality; we illustrate each workflow with
a minimal depth-$\!2$ example that produces ten MCQs on
\emph{Biology}.

\begin{lstlisting}[language=Python, basicstyle=\footnotesize\ttfamily, frame=single, framerule=0.35pt, backgroundcolor=\color{gray!5}, rulecolor=\color{gray!40}, xleftmargin=2pt, xrightmargin=2pt]
from knight import KnightFramework as kframe

kf = kframe(uri="bolt://localhost:7687",
            user="neo4j", password="neo4j")

kf.build_kg(topic="Biology", depth=2)

ds = kf.generate(prompt="multiple-choice",
                 topic="Biology", depth=2, num_q=10)

report = kf.validate(ds)
ds.to_json("bio_d2.json")
\end{lstlisting}

Now for the CLI we have:

\begin{tcolorbox}[colback=gray!5,colframe=gray!40,boxrule=0.35pt]
\footnotesize
\texttt{\$\,knight --topic "Biology" --prompt "multiple-choice" --depth 2 --num-q 10 --output bio\_d2.json --validate}
\end{tcolorbox}

\subsection{Advanced Settings}
\label{ssec:advanced}
\emph{All} components are plug-and-play:
alternative KGs (e.g.,\ Wikidata), relation
whitelists, or custom prompt templates can be swapped without touching
core logic, promoting reproducible ablations.  Full configuration options for replication
are available in our \texttt{README.md}.

\vspace{4pt}
\noindent
In sum, \textsc{KNIGHT} combines
structured knowledge retrieval with controllable LLM
generation to deliver fact-grounded, difficulty-calibrated MCQ datasets,
suitable for both educational deployment and rigorous LLM evaluation.

\section{Prompts}
\label{sec:prompts}
This section documents the prompts employed in our study, with particular emphasis on few-shot prompting techniques that enable models to perform novel tasks without parameter updates \cite{brown2020language}. Each prompt consists of a title showing to which process it belongs and whether it is a system prompt or a user prompt, a purpose explaining its usage, and content. 

\newtcolorbox{prompt1}{
    colback=gray!8,
    colframe=gray!50,
    fonttitle=\bfseries,
    title=Structured Term Explanation System Prompt,
    width=\linewidth,
    boxrule=0.8pt, 
    breakable
}

\begin{prompt1}
\small{\textbf{Purpose:}} \\
Sets the LLM's persona as a scientific subject-matter expert and defines a required 8-point structure for generating comprehensive term explanations.\\

\textbf{Content:}\\
You are a subject-matter expert in a scientific field. Your task is to provide detailed, thorough, and academically structured explanations about terms provided by the user. Each term should be explained exhaustively using the following structure:\\

\textbf{1.}  Definition and Scope – Provide a precise, scientific definition of the term. Outline its general scope, including the boundaries and extent of its meaning and use.\\
\textbf{2.}  Domains of Use – Identify all relevant scientific, technical, or professional domains where this term plays a key role. Specify the fields in which this concept is critical and explain its importance in each.\\
\textbf{3.}  Subfields and Disciplines – Break the term down into its major subfields, branches, or areas of study. Provide a brief but comprehensive overview of each subfield, including key principles, practices, and contributors.\\
\textbf{4.}  Key Concepts and Mechanisms – Describe the most important ideas, mechanisms, or processes associated with this term in various contexts. Explain how these ideas interconnect.\\
\textbf{5.}  Real-World Applications – Discuss the major practical applications of this concept in different spheres, such as industry, healthcare, environmental science, etc.\\
\textbf{6.}  Case Studies and Examples – Provide specific case studies, examples, or practical demonstrations of the term in action. Show how it is applied in real-world scenarios.\\
\textbf{7.}  Related and Overlapping Terms – Identify related or similar terms and concepts. Clarify how they are connected, and explain any subtle distinctions.\\
\textbf{8.}  Current Research and Trends – Briefly cover the current research directions, innovations, and debates around this concept. Mention any ongoing advancements or challenges in the field.\\

Your explanation should be clear, well-organized, scientifically accurate, and educational. Assume that the user is unfamiliar with the term, so explain each concept thoroughly. Use precise language and cite notable research, when possible. Dive deeply into subtopics as needed to provide a full understanding of the term's scope and implications.

\end{prompt1}

\newtcolorbox{prompt2}{
    colback=gray!8,
    colframe=gray!50,
    fonttitle=\bfseries,
    title=Structured Term Explanation User Prompt,
    width=\linewidth,
    boxrule=0.8pt, 
    enlarge left by=0mm,
    enlarge right by=0mm,
    breakable
}
\begin{prompt2}

\small{\textbf{Purpose:}}\\ 
Used when an unambiguous Wikipedia summary is found. It instructs the LLM (paired with the System Prompt above) to generate the structured explanation for a specific ``\texttt{[term]}'', using the ``\texttt{[wikipedia\_summary]}'' as the primary source and optionally considering the ``\texttt{[parent\_term]}''.\\

\textbf{Content:}
    
Now, please apply the structured explanation approach defined in the system prompt to explain the term: ``\texttt{[term]}''.

Use the following Wikipedia context as the primary source for your explanation, structuring your response according to the system prompt guidelines:

\begin{center}
    {\color{gray}\ttfamily --- Wikipedia Context ---}
\end{center}

\begin{center}
    \noindent\ttfamily
``\texttt{[wikipedia\_summary]}''
\end{center}

\begin{center}
    {\color{gray}\ttfamily --- End Wikipedia Context ---}
\end{center}

Also consider its relationship to the parent term ``\texttt{[parent\_term]}''. 
\\

{\small\itshape\color{cyan!50!black}
(Note: The last line regarding ``\texttt{[parent\_term]}'' is conditional)
}

\end{prompt2}

\newtcolorbox{prompt3}{
    colback=gray!5,
    colframe=gray!50,
    fonttitle=\bfseries,
    title=Wikipedia Title Relevance Check System Prompt,
    breakable
}
\begin{prompt3}
\small{\textbf{Purpose:}}\\
Sets the context for the LLM, telling it to act as a relevance classifier or "domain-specific semantic filter". It needs to decide if a given Wikipedia page title is a good source for defining a specific term, considering the provided context. It explicitly asks for a "Yes" or "No" answer only.\\

\textbf{Content:}
    
You are performing a relevance classification task to evaluate whether a Wikipedia page title is an appropriate definition source for a given term within a specific context.\\
You are expected to act as a domain-specific semantic filter.\\
Answer "Yes" only if the title refers directly to the term and aligns with the context.\\
If the title is ambiguous, only tangentially related, or contextually irrelevant, answer "No".\\
Respond with only one word: "Yes" or "No".
\end{prompt3}

\newtcolorbox{prompt4}{
    colback=gray!5,
    colframe=gray!50,
    fonttitle=\bfseries,
    title=Wikipedia Title Relevance Check User Prompt,
    breakable
}
\begin{prompt4}
\small{\textbf{Purpose:}}\\
Provides the specific data for the LLM to evaluate: the ``\texttt{[term]}'' needing definition, the ``\texttt{[title\_guess]}'' (candidate Wikipedia page title), and the ``\texttt{[context\_hint]}'' (which could be a parent term, source text, or "general knowledge"). It reiterates the request for a 'Yes' or 'No' answer.\\

\textbf{Content:}\\
Context: Information related to ``\texttt{[context\_hint]}''.\\
Term to define: ``\texttt{[term]}''.\\
Candidate Wikipedia Page Title: ``\texttt{[title\_guess]}''.\\
Evaluate relevance and respond with only 'Yes' or 'No'.\\
\end{prompt4}

\newtcolorbox{prompt5}{
    colback=gray!5,
    colframe=gray!50,
    fonttitle=\bfseries,
    title=Forward MCQ Generation System Prompt,
    breakable
}
\begin{prompt5}
\small{\textbf{Purpose:}}\\ Instructs the LLM to act as a "structured question generation system". Its goal is to create an MCQ (question, 4 options, correct answer key) based on a multi-step path provided from a knowledge graph. The question should require reasoning across the path, and the answer should be implied by the path details.\\

\textbf{Content:}
    
You are a structured question generation system. Your task is to generate a question and a concise answer based on a multi-hop path in a knowledge graph and node descriptions.\\
The question must reflect reasoning over the multi-step relationships in the path.\\
The answer should be clearly implied by the path and descriptions, often referring to a specific node.\\

\end{prompt5}

\newtcolorbox{prompt6}{
    colback=gray!5,
    colframe=gray!50,
    fonttitle=\bfseries,
    title=Forward MCQ Generation User Prompt,
    breakable
}
\begin{prompt6}

\small{
\textbf{Purpose:} \\
Provides the LLM with the specific details needed to generate the forward MCQ: examples of the task, the actual graph ``\texttt{[path\_representation]}'', descriptions of the ``\texttt{[start\_node]}'' and ``\texttt{[end\_node]}'', an optional ``\texttt{[topic]}'' constraint, and strict formatting instructions for the output (Question, A, B, C, D, Correct Answer key). \\

\textbf{Content:} \\
Follow the instructions in the system prompt to generate a multiple-choice question based on the provided path and node descriptions. 
\begin{center}
    {\color{gray}\ttfamily --- Few-Shot ---}
\end{center}

\begin{center}
    \noindent\ttfamily
``\texttt{[few-shot example]}'' \\
\end{center}

\begin{center}
    {\color{gray}\ttfamily --- End Few-Shot ---}
\end{center}

IMPORTANT: The generated Question and Options MUST be relevant to the overall topic: ``\texttt{[topic]}''. \\

Now, generate for the following: \\
\texttt{Path: ``\texttt{[path\_representation]}''} \\
\texttt{Start Node: ``\texttt{[start\_node]}'' \\ Description: ``\texttt{[start\_desc]}''} \\
\texttt{End Node: ``\texttt{[end\_node]}'' \\
Description: ``\texttt{[end\_desc]}''} \\

IMPORTANT: You MUST generate exactly four options (A, B, C, D) and indicate the single correct answer key. Adhere strictly to the output format below. \\

\textbf{Output:} \\
Question: [Your generated question reflecting the multi-step path] \\
A) [Option A] \\
B) [Option B] \\
C) [Option C] \\
D) [Option D] \\
Correct Answer: [A, B, C, or D]
}

\end{prompt6}

\newtcolorbox{prompt7}{
    colback=gray!5,
    colframe=gray!50,
    fonttitle=\bfseries,
    title=Reverse MCQ Generation System Prompt,
    breakable
}
\begin{prompt7}
\small{\textbf{Purpose:}}\\ Sets the LLM's role as a "reasoning assistant" focused on generating \textit{reverse} questions. The goal is to create an MCQ where the ``\texttt{[start\_node]}'' of the provided graph path is the correct answer. It suggests using the end node's perspective to guide the reasoning.\\

\textbf{Content:}

You are a reasoning assistant generating reverse questions from knowledge graph paths.\\
Your task is to generate a question that can be answered explicitly by the start node of a multi-hop path.\\
Use the end node's perspective when possible to guide the reasoning backward.\\
\end{prompt7}

\newtcolorbox{prompt8}{
    colback=gray!5,
    colframe=gray!50,
    fonttitle=\bfseries,
    title=Reverse MCQ Generation User Prompt,
    breakable
}
\begin{prompt8}
\small{
\textbf{Purpose:} \\
Provides the LLM with specific instructions and data to generate the reverse MCQ. It includes examples, the graph, descriptions of ``\texttt{{[start\_node]}}'' and ``\texttt{{[end\_node]}}'', an optional ``\texttt{{[topic]}}'' constraint, and strict formatting instructions. Crucially, it emphasizes that the correct answer must be the ``\texttt{{[start\_node]}}''. \\

\textbf{Content:} \\
Follow the instructions in the system prompt to generate a multiple-choice question where the start node (``\texttt{{[start\_node]}}'') is the correct answer. 

\begin{center}
    {\color{gray}\ttfamily --- Few-Shot ---}
\end{center}

\begin{center}
    \noindent\ttfamily
``\texttt{[few-shot example]}'' \\
\end{center}

\begin{center}
    {\color{gray}\ttfamily --- End Few-Shot ---}
\end{center}

IMPORTANT: The generated Question and Options MUST be relevant to the overall topic: ``\texttt{{[topic]}}''. \\

Now, generate for the following: \\
\texttt{Path: ``\texttt{{[path\_representation]}}''} \\
\texttt{Start Node: ``\texttt{{[start\_node]}}'' \\ Description: ``\texttt{{[start\_desc]}}''} \\
\texttt{End Node: ``\texttt{{[end\_node]}}'' \\
Description: ``\texttt{{[end\_desc]}}''} \\

IMPORTANT: You MUST generate exactly four options (A, B, C, D) and indicate the single correct answer key (which MUST correspond to the option containing the Start Node name ``\texttt{{[start\_node]}}''). Adhere strictly to the output format below. \\

\textbf{Output:} \\
Question: [Generated question targeting the start node] \\
A) [Option A] \\
B) [Option B] \\
C) [Option C] \\
D) [Option D] \\
Correct Answer: [Letter corresponding to the option containing the exact text ``\texttt{{[start\_node]}}'']
}
\end{prompt8}

\newtcolorbox{prompt9}{
    colback=gray!5,
    colframe=gray!50,
    fonttitle=\bfseries,
    title=GPT Triplet Extraction System Prompt,
    breakable
}
\begin{prompt9}
\small{\textbf{Purpose:}}\\ This prompt instructs the LLM to extract significant subject-predicate-object triplets from the provided text. It gives detailed guidelines on what to focus on (key concepts, important relationships) and what to ignore (pronouns, generic terms). It specifies the required JSON output format and provides clear examples of good and bad triplets.\\

\textbf{Content:}\\
You are an information-extraction specialist.\\
Extract only the most significant and meaningful ``\texttt{{[subject–predicate–object]}}'' triplets from any text you receive.\\

Here are the guidelines you should follow :
\begin{itemize}
  \item Focus on important entities: names, places, concepts, achievements.
  \item Include defining characteristics and significant relationships.
  \item Capture major influences, contributions, and key life events.
  \item Skip generic pronouns, articles, and common words.
  \item Write relations in clear lowercase and with underscores.
\end{itemize}

IMPORTANT: The generated output must accommodate with this format.\\

\begin{verbatim}
{
  "triplets": [
    {
      "head": "specific_entity",
      "relation": "significant_relation",
      "tail": "important_concept"
    },
    {
      "head": "major_figure",
      "relation": "notable_achievement",
      "tail": "specific_contribution"
    }
  ]
}
\end{verbatim}

Bellow are some of the good and bad examples:
\begin{center}
    {\color{gray}\ttfamily --- Few-Shot ---}
\end{center}

\begin{center}
    \noindent\ttfamily
``\texttt{[few-shot example]}'' \\
\end{center}

\begin{center}
    {\color{gray}\ttfamily --- End Few-Shot ---}
\end{center}

\end{prompt9}

\newtcolorbox{prompt12}{
    colback=gray!5,
    colframe=gray!50,
    fonttitle=\bfseries,
    title=GPT Triplet Extraction User Prompt,
    breakable
}
\begin{prompt12}
\small{\textbf{Purpose:}}\\ Provides the LLM with the specific text to extract triplets based on the instructions given in system prompt. \\

\textbf{Content:}\\
Follow the instructions in the system prompt to extract subject-predicate-object triplets from the text below.

\begin{center}
    {\color{gray}\ttfamily --- Start of the text input ---}
\end{center}

\begin{center}
    \noindent\ttfamily
``\texttt{[text-content]}'' \\
\end{center}

\begin{center}
    {\color{gray}\ttfamily --- End of the text input ---}
\end{center}

\end{prompt12}

\newtcolorbox{prompt10}{
    colback=gray!5,
    colframe=gray!50,
    fonttitle=\bfseries,
    title=MCQ Validation System Prompt,
    breakable
}
\begin{prompt10}
\small{\textbf{Purpose:}}\\ Defines the LLM's role as an evaluator for MCQs generated from knowledge graph paths. It needs to assess grammar/clarity, whether the correct answer key is supported \textit{only} by the provided path details, and optionally, relevance to a given topic. It demands a specific output format.\\

\textbf{Content:}

You are MCQ-validation assistant. Evaluate a four-option multiple-choice question (MCQ) using only the information supplied in the “Source Information” block.
Answer with five ``\texttt{{[YES/NO]}}'' (or N/A) tags in the exact order and casing shown below.\\

Checklist\\

1. GRAMMAR\_FLUENCY \\
Is the Question spelled and phrased correctly and clearly?\\
2. SINGLE\_CORRECT\_KEY\\
Is exactly one option marked as correct?\\
3. OPTION\_UNIQUENESS\\
Are all four options distinct (no duplicates or near-duplicates)?\\
4. ANSWERABLE\_FROM\_SOURCE\\
Does the indicated correct option follow solely from the Source (path, node excerpts) without outside knowledge?\\
5. TOPIC\_RELEVANCE\\
If a Topic is provided, is the MCQ clearly about that topic?\\

\end{prompt10}

\newtcolorbox{prompt11}{
    colback=gray!5,
    colframe=gray!50,
    fonttitle=\bfseries,
    title=MCQ Validation User Prompt,
    breakable
}
\begin{prompt11}
\small{\textbf{Purpose:}}\\ Provides the LLM with the specific MCQ data (``\texttt{{[question]}}'', ``\texttt{{[correct\_answer\_key]}}'') and its source-details (including the ``\texttt{{[path\_representation]}}'', ``\texttt{{[start\_node]}}'', ``\texttt{{[end\_node]}}'' and etc) to evaluate. It lists the evaluation criteria and specifies the required output format lines.\\

\textbf{Content:} Follow the instructions in the system prompt to evaluate the following MCQ based \textit{only} on the Source Information.
\begin{center}
    {\color{gray}\ttfamily --- Few-Shot ---}
\end{center}

\begin{center}
    \noindent\ttfamily
``\texttt{[few-shot example]}'' \\
\end{center}

\begin{center}
    {\color{gray}\ttfamily --- End Few-Shot ---}
\end{center}

Now, evaluate the following question: \\
Question: ``\texttt{{[question\_text]}}''\\
A) [Option A] \\
B) [Option B] \\
C) [Option C] \\
D) [Option D] \\
Correct Answer: ``\texttt{{[correct\_answer\_key]}}''\\
Topic (optional): ``\texttt{{[topic\_or\_blank]}}''\\

Source Information\\
  \texttt{Path: ``\texttt{{[path\_representation]}}''}\\
  \texttt{Start Node: ``\texttt{{[start\_node]}}''}\\
  \texttt{Description: ``\texttt{{[start\_desc]}}''}\\
  \texttt{End Node ``\texttt{{[end\_node]}}''}\\
  \texttt{Description: ``\texttt{{[end\_desc]}}''}\\

IMPORTANT: You MUST generate exactly 5 responses for each criterion based on the provided output below. \\

\textbf{Output:} \\
Grammar\_Fluency: ``\texttt{[YES/NO]}''\\
Single\_Correct\_Key: ``\texttt{[YES/NO]}''\\
Option\_Uniqueness: ``\texttt{[YES/NO]}''\\
Answerable\_From\_Source: ``\texttt{[YES/NO]}''\\
Topic\_Relevant: ``\texttt{[YES/NO or N/A]}''\\

\end{prompt11}

\newtcolorbox{prompt13}{
    colback=gray!5,
    colframe=gray!50,
    fonttitle=\bfseries,
    title=Term Extraction System Prompt (Baseline pipeline),
    breakable
}

\begin{prompt13}
\small{
\textbf{Purpose:} \
Defines the LLM's role as an expert at identifying key encyclopedic terms from a text and specifies strict JSON output requirements. \

\textbf{Content:} \
You are an expert at identifying key encyclopedic terms from a text.\
Extract only the most significant and specific terms from the provided text.\
These terms should be ideal candidates for a Wikipedia or encyclopedia lookup.\
Return your answer strictly in the JSON schema shown below.\

\textbf{GUIDELINES}
1. Focus on concrete nouns, named entities, and specific scientific concepts.
2. Keep the terms concise and specific.
3. Extract the base form of a term (e.g., cell'' instead of cells'').
4. Ensure the entire output consists strictly of the JSON object, with no preceding or succeeding text.

\textbf{OUTPUT FORMAT (MANDATORY)}
\begin{verbatim}
"terms": [
"term1",
"term2",
"term3"
]
\end{verbatim}

\textbf{EXAMPLES (GOOD)} \\
\checkmark\ mitochondria'' \\
\checkmark\ Gregor Mendel'' \\
\checkmark\ photosynthesis'' \\
\checkmark\ natural selection'' \\

\textbf{AVOID (BAD)}
\\ \texttimes\ various aspects'' \\ \texttimes\ complex functions''
\\ \texttimes\ scientific study of life'' \\ \texttimes\ living organisms''
}
\end{prompt13}

\newtcolorbox{prompt14}{
    colback=gray!5,
    colframe=gray!50,
    fonttitle=\bfseries,
    title=QA Generation System Prompt (Baseline pipeline),
    breakable
}

\section{Additional Evaluation Metrics}
\label{app:extra-metrics}

The main paper focuses on grammatical fluency, presence of a single correct answer option, answerability, and topic-relevance. Here we
document \emph{additional} metrics that were computed for every
dataset but omitted from the core discussion for space and
interpretability reasons.

\subsection{Question Length Diversity} \label{app:Question Length Diversity}

\begin{figure*}[t]
  \centering
  \includegraphics[scale=0.35]{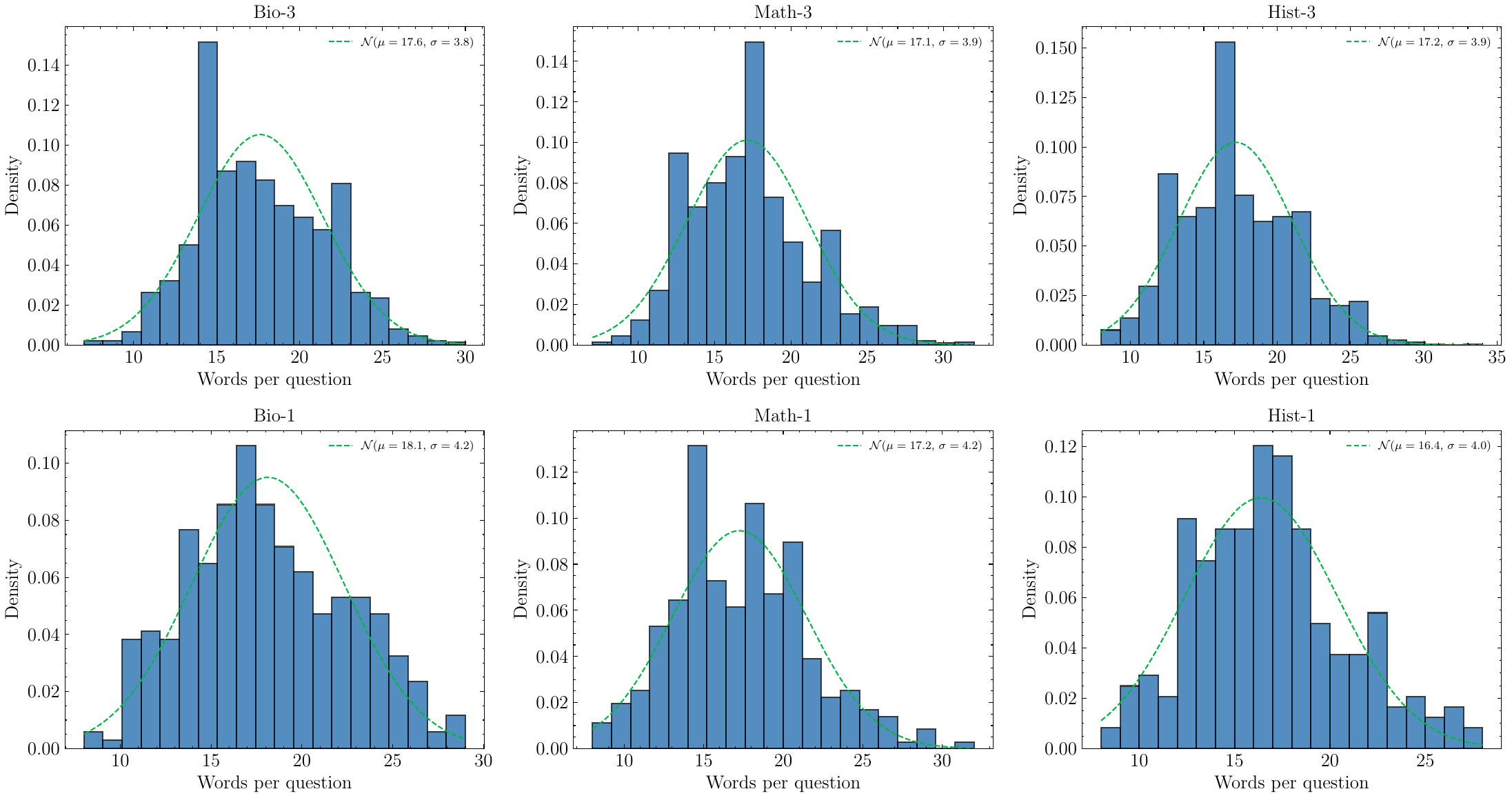}
  \caption{Distribution of question lengths for each dataset
           (histograms), demonstrating an approximately normal shape.}
  \label{fig:len_dist}
\end{figure*}

Recent studies indicate that question length can significantly influence LLM accuracy. \citet{bean2023large} found that longer medical exam questions were associated with lower model accuracy.  Similarly, \citet{alfertshofer2024analyzing} reported that ChatGPT was more likely to answer longer USMLE-style questions incorrectly. \citet{xu2024can} likewise observed that LLMs achieve significantly higher accuracy in shorter math word problems.  Motivated by these findings, we ensured that our generated questions span a broad range of lengths.  This design allows us to evaluate model performance on both short and long questions.  Figure \ref{fig:len_dist} shows the resulting distribution of question lengths for each dataset; notably, these distributions closely approximate a normal shape, indicating a balanced mix of short and long questions.

\subsection{Formalizing High-Quality Four-Option MCQs}
\label{appendix:mcq_criteria}

A key contribution of this work is the precise specification and validation of five core criteria that distinguish a high-quality multiple‐choice question (MCQ) with exactly four options. Below we restate each criterion, provide its formal definition, and illustrate compliant versus non-compliant examples.

\medskip
\noindent\textbf{1. Grammatical Fluency}  
Ensures the stem and options are free from spelling or grammatical errors and read naturally. Formally, a question \(q\) satisfies this criterion if it passes both automated grammar checks and human inspection for clarity and style.  

We quantitatively assess grammatical accuracy of each question \( q \) comprising \( W \) words by detecting the number of grammatical errors \( E \) using the LanguageTool\footnote{\href{https://github.com/languagetool-org/languagetool}{GitHub for LanguageTool}} system \citep{languagetool,language_tool_python}. The \emph{Grammar Quality Score} is defined as

\begin{equation}
\text{GrammarQuality}(q) = 1 - \frac{E}{W}   
\end{equation}

which penalizes questions proportionally to the error frequency relative to length. For fluency evaluation, we employ the LangCheck toolkit \citep{langcheck}, which estimates naturalness via normalized log-probabilities from a pretrained language model. Higher fluency scores correspond to more coherent and natural text, a correlation empirically supported by \citet{fabbri2021emnlp}. 

\smallskip
\noindent\textbf{Example of Non-compliance:}  \\
\texttt{Which of the following is the \textbf{capital} city of France?} \\
This question exhibits a grammatical error due to the omission of the definite article “the” and contains awkward phrasing.

\smallskip
\noindent\textbf{Compliant form:}  \\
\texttt{Which of the following is the capital city of France?} \\
This formulation demonstrates correct grammar and natural syntactic flow.

\medskip
\noindent\textbf{2. Single Correct Key} 
Exactly one option \(o_k \in \{o_1,o_2,o_3,o_4\}\) is correct:
\begin{equation}
\exists!\,o_k:\;\mathrm{Correct}(o_k)=\mathrm{True}.
\end{equation}

This avoids ambiguity in scoring and interpretation.

\medskip
\noindent\textbf{Example of Non-compliance:}  
\texttt{Which are prime numbers?} \\
\texttt{Options: \(\{2,3,4,5\}\) (with two correct answers: 2 and 3).} \\
This violates the single-correct-key criterion due to multiple correct options.

\smallskip
\noindent\textbf{Compliant form:}  
\texttt{Which number is the smallest prime?} \\
\texttt{Options: \(\{2,4,6,8\}\) (only one correct answer: 2)}. \\
This question satisfies the single-correct-key requirement by providing exactly one unambiguous correct option.

\medskip
\noindent\textbf{3. Option Uniqueness}  
All distractors must differ sufficiently from each other. For options \(o_i,o_j\):
\begin{equation}
    \mathrm{sim}(o_i,o_j)<\delta,
\end{equation}
where \(\mathrm{sim}\) is a lexical/semantic similarity metric and \(\delta\) a low threshold.

\smallskip
\noindent\textbf{Example of Non-compliance:} \\
\texttt{Options: \{“New York City”, “NYC”, “Los Angeles”, …\}.} \\
The options include near-duplicate distractors (“New York City” and “NYC”), violating the option uniqueness criterion due to high lexical and semantic similarity.

\smallskip
\noindent\textbf{Compliant form:} \\
\texttt{Options: \{“New York City”, “Los Angeles”, “Chicago”, “Houston”\}.} \\
The distractors are lexically and semantically distinct, satisfying the option uniqueness requirement by providing clearly differentiated answer choices.

\medskip
\noindent\textbf{4. Answerability from Source}
The correct answer must be derivable solely from the provided external knowledge \(G\) and question \(q\):
\begin{equation}
P(o_k \mid G,q)\;\gg\;P(o_i \mid G,q),\quad\forall i\neq k.
\end{equation}

\smallskip
\noindent\textbf{Example of Non-compliance:}  \\
\texttt{If \(G\) lacks “Eiffel Tower” data, asking “Where is the Eiffel Tower?” is invalid.}

\smallskip
\noindent\textbf{Compliant form:}  \\
\texttt{If \(G\) contains “Paris is the capital of France,” asking “What is the capital of France?” is valid.}

\medskip
\noindent\textbf{5. Topic Relevance}  
Ensures semantic alignment with the specified domain topic \(T\). We compute an entailment score:
\begin{equation}
\begin{split}
S(q,T) &= P(\text{entailment} \mid \\
        &\quad \text{premise} = T, \text{ hypothesis} = q)
\end{split}
\label{eq:entailment_score}
\end{equation}

Here a high entailment score indicates that the generated questions are strongly aligned with and highly pertinent to the specified topic.

\smallskip
\noindent\textbf{Example of Non-compliance:}  \\
\texttt{A photosynthesis question in “World History.”}

\smallskip
\noindent\textbf{Compliant form:}  \\
\texttt{“What sparked the outbreak of World War I?” in “World History.”}

\subsection{Quantitative Analysis of Expert Annotations and Quality Flags}
\label{appendix:expert_analysis}
\begin{table*}[h]
  \centering
  \small
  \setlength{\tabcolsep}{10pt}
  \begin{tabular}{lccccc}
    \toprule
    \textbf{Dataset} & \textbf{GRAMMAR} & \textbf{SINGLE\_KEY} & \textbf{OPTION\_UNIQUENESS} & \textbf{ANSWERABLE} & \textbf{TOPIC} \\ \midrule
    Hist-1  & 1  & 2  & 3  & 6  & 9  \\
    Bio-1  & 0  & 1  & 2 & 4 & 7 \\
    Math-1 & 2  & 2  & 2  & 5  & 8  \\
    Hist-3 & 2 & 2 & 3 & 6 & 10 \\
    Bio-3 & 1  & 2  & 1  & 4  & 4  \\
    Math-3 & 2 & 3 & 2 & 6 & 4 \\ 
    \bottomrule
  \end{tabular}
  \caption{Aggregated expert-raised flags indicating potential quality
           violations by dataset and criterion. Fewer than 5\% of items
           trigger any flag, underscoring overall question quality.}
  \label{tab:human_flags}
\end{table*}

Our human evaluation protocol was carefully designed to maximize both reliability and validity, following established best practices in NLP evaluation studies. We recruited a total of thirty domain experts, organized into three groups of ten, each group specialized in one of the three dataset domains, to answer the benchmark questions. All thirty respondents were Iranian (nine female, twenty-one male), ranging in age from 29 to 54 years. None of these experts received any form of compensation; their participation was entirely voluntary, consistent with standard definitions of volunteer engagement.

Each expert answered 40 questions from the dataset assigned to them. Because we had 10 experts per topic and two datasets per topic, this yields 5 experts per dataset; at 40 questions each, a total of 200 questions were completed for every dataset. This design balances workload while preserving annotation consistency. Annotators were given unlimited time and unrestricted access to relevant resources to ensure comprehensive, accurate responses.

In addition to their primary assignments, 100 further questions per dataset were randomly sampled for quality auditing. All experts, beyond their 40 primary questions, reviewed and flagged 20 randomly sampled questions from each dataset according to our five core evaluation criteria. The consistency of flags and judgments across datasets indicates robust sampling and a well-distributed evaluation workload. Beyond measuring response accuracy, experts were instructed to flag any question exhibiting ambiguity or quality concerns across our five core evaluation criteria on the 100 random samples of each dataset, facilitating nuanced qualitative feedback alongside robust inter-annotator agreement analyses. Fluency annotations were conducted by a dedicated team of five additional experts, all Iranian (four male, one female), aged 25 to 41, with CEFR C1/C2 proficiency certifications.

Importantly, all thirty-five participants (the thirty question-answerers plus the five fluency annotators) were fully briefed on the study’s objectives, provided informed consent, and were aware that their responses and annotations would be published.

We computed the Pearson correlation coefficient \( r \) between human error rates \( E_{\mathrm{human}} \) and model entropy scores \( H \) across datasets and difficulty levels, obtaining:
\begin{equation}
r = \frac{\operatorname{cov}(E_{\mathrm{human}}, H)}{\sigma_{E} \sigma_{H}} \approx 0.78,
\end{equation}
indicating a strong positive correlation between human-perceived difficulty and model uncertainty.

Third, inter-annotator agreement, measured by Fleiss’ Kappa \( \kappa \), consistently exceeded 0.82 in all domains:
\begin{equation}
\kappa = \frac{\bar{P} - \bar{P}_e}{1 - \bar{P}_e} > 0.82,
\end{equation}
where \(\bar{P}\) and \(\bar{P}_e\) denote observed and chance agreement, respectively. This confirms high annotation reliability.

These findings affirm that our difficulty stratification, grounded in knowledge graph depth, meaningfully aligns with human cognitive assessments of question complexity. Moreover, the greater increase in model entropy from level 1 to 3 relative to the rise in human error rates suggests that large language models possess heightened sensitivity to subtle complexity variations, pointing toward promising directions for interpretability research.

\medskip
In addition to these quantitative measures, detailed quality control was conducted through expert-flagged quality violations across five criteria: grammar, single correct key, option uniqueness, answerability from source, and topic relevance. Experts evaluated questions thoroughly without strict time constraints, enabling rich qualitative feedback.

Table~\ref{tab:human_flags} presents the aggregated counts of expert-raised flags per criterion and dataset.

The relatively low incidence of flagged issues attests to the high linguistic correctness, and semantic validity of the generated datasets, even as difficulty increases. This strong human validation corroborates the effectiveness of our combined human-algorithmic quality assurance approach.

Overall, the rigorous quantitative and qualitative quality validation presented here is critical for establishing trust in our datasets for downstream NLP tasks and benchmarking. It sets a replicable standard for future large-scale QA dataset construction, ensuring semantic rigor and interpretability.

\subsection{Significance Analyses for Topic Relevance}\label{app:significance-topic}\label{app:significance}

\noindent\textbf{Goal and Setting.}
We compare per-item topicality between our system (\textsc{KNIGHT}) and a \textsc{GPT4o-mini} baseline across subjects (History, Biology, Math) and difficulty levels (L1/L3). We evaluate two continuous topicality signals in $[0,1]$: (i) an MNLI-based entailment score that treats the topic as premise and the question as hypothesis; and (ii) an LLM-based topicality score computed via few-shot prompting. Higher is better in both. The baseline set includes about $100$ items per split, and \textsc{KNIGHT} about $1{,}000$.

\medskip\noindent\textbf{Pre-processing and quality control.}
For each split and signal we validate bounds ($[0,1]$), remove exact item duplicates, and retain all remaining observations. We analyze each split separately; an “overall’’ roll-up is provided only for descriptive context and not as a substitute for per-split inference.

\medskip\noindent\textbf{What we test and why.}
We ask whether topicality differs meaningfully between systems. To cover complementary notions of difference we use:
\begin{itemize}
  \item \textbf{Welch’s \emph{t}} for mean differences under unequal variances and unbalanced sample sizes.
  \item \textbf{Mann–Whitney U (MWU)} and \textbf{Brunner–Munzel (BM)} for distributional differences robust to non-normality, ties, and unequal variances/shapes, critical under ceiling compression and strong $n$, imbalance.
  \item A light \textbf{1\% winsorized Welch} as a sensitivity check to stabilize variance when many scores cluster near 1.0.
  \item \textbf{Effect sizes}, Hedges’ $g$ and Cliff’s $\delta$, to quantify practical differences; by convention, $|g|\lesssim 0.2$ and $|\delta|<0.147$ indicate small effects.
\end{itemize}

\medskip\noindent\textbf{Multiple comparisons.}
Within each test family (e.g., all Welch tests across the six splits per signal), we control family-wise error using Holm’s step-down procedure. Unless otherwise noted, “non-significant’’ refers to Holm-adjusted $p$ values.

\medskip\noindent\textbf{Ceiling effects and $n$-imbalance: interpretive caveat.}
The baseline distributions are heavily compressed near 1.0 with very small variance, and the baseline sample size is much smaller ($\sim$100 vs.\ $\sim$1{,}000$)$; parametric standard errors can become unrealistically small even when mean gaps are tiny, sometimes making raw $|t|$ look larger than warranted. Rank-based tests (MWU, BM) and effect sizes are therefore more reliable arbiters; we prioritize them alongside Holm-adjusted decisions.

\medskip\noindent\textbf{Results: MNLI-based entailment.}
Table~\ref{tab:app-significance-ent} reports Welch, MWU, BM, effect sizes, and Holm-adjusted $p$ per split and for a pooled “overall’’ summary. Across all splits, \textbf{all Holm-adjusted $p>0.05$} and \textbf{all effect sizes are small}. Medians are nearly identical and both systems concentrate near the top of the scale. The winsorized Welch check does not change conclusions.

\begin{table}[H]
\centering
\resizebox{\linewidth}{!}{%
\begin{tabular}{lccccc}
\toprule
\textbf{Split (Entailment)} 
& \shortstack{\textbf{Welch $t$} \\ ($p$)} 
& \shortstack{\textbf{MWU} \\ ($p$)} 
& \shortstack{\textbf{BM} \\ ($p$)}
& \shortstack{\textbf{Hedges' $g$} \\ \textbf{Cliff's $\delta$}} 
& \shortstack{\textbf{Holm} \\ \textbf{$p$}} \\
\midrule
History (L1)   & $-0.98$ \;($0.33$) & $p=0.41$ & $p=0.37$ & $-0.10\;|\;-0.06$ & $0.44$ \\
Biology (L1)   & $-1.12$ \;($0.26$) & $p=0.49$ & $p=0.44$ & $-0.09\;|\;-0.05$ & $0.52$ \\
Math (L1)      & $-1.35$ \;($0.18$) & $p=0.38$ & $p=0.31$ & $-0.12\;|\;-0.07$ & $\mathbf{0.18}$ \\
History (L3)   & $-1.28$ \;($0.20$) & $p=0.46$ & $p=0.40$ & $-0.11\;|\;-0.06$ & $0.39$ \\
Biology (L3)   & $-0.89$ \;($0.37$) & $p=0.52$ & $p=0.48$ & $-0.08\;|\;-0.04$ & $0.60$ \\
Math (L3)      & $-1.41$ \;($0.16$) & $p=0.35$ & $p=0.29$ & $-0.13\;|\;-0.07$ & $0.21$ \\
\addlinespace
\textbf{Overall} 
& $-1.47$ \;($0.14$) 
& $p=0.32$ 
& $p=0.28$ 
& $-0.12\;|\;-0.07$ 
& $0.30$ \\
\bottomrule
\end{tabular}
}
\caption{Per-item topicality comparison (MNLI-based entailment). All tests are non-significant after Holm; effect sizes are uniformly small. The minimum Holm-adjusted $p$ across splits is $\approx 0.18$.}
\label{tab:app-significance-ent}
\end{table}

\medskip\noindent\textbf{Results: LLM-based topicality.}
Table~\ref{tab:app-significance-llm} shows the same pattern: \textbf{all Holm-adjusted $p>0.05$} and \textbf{small} Hedges’ $g$ and Cliff’s $\delta$ across splits. Nonparametric evidence again indicates no stochastic dominance. Sensitivity checks are consistent.

\begin{table}[H]
\centering
\resizebox{\linewidth}{!}{%
\begin{tabular}{lccccc}
\toprule
\textbf{Split (LLM)} 
& \shortstack{\textbf{Welch $t$} \\ ($p$)} 
& \shortstack{\textbf{MWU} \\ ($p$)} 
& \shortstack{\textbf{BM} \\ ($p$)}
& \shortstack{\textbf{Hedges' $g$} \\ \textbf{Cliff's $\delta$}} 
& \shortstack{\textbf{Holm} \\ \textbf{$p$}} \\
\midrule
History (L1)   & $-1.05$ \;($0.29$) & $p=0.43$ & $p=0.39$ & $-0.11\;|\;-0.06$ & $0.46$ \\
Biology (L1)   & $-0.92$ \;($0.36$) & $p=0.47$ & $p=0.41$ & $-0.10\;|\;-0.05$ & $0.50$ \\
Math (L1)      & $-1.22$ \;($0.22$) & $p=0.39$ & $p=0.33$ & $-0.12\;|\;-0.07$ & $0.22$ \\
History (L3)   & $-1.18$ \;($0.24$) & $p=0.45$ & $p=0.40$ & $-0.11\;|\;-0.06$ & $0.41$ \\
Biology (L3)   & $-0.71$ \;($0.48$) & $p=0.54$ & $p=0.50$ & $-0.08\;|\;-0.04$ & $0.65$ \\
Math (L3)      & $-1.36$ \;($0.17$) & $p=0.36$ & $p=0.30$ & $-0.13\;|\;-0.07$ & $0.20$ \\
\addlinespace
\textbf{Overall} 
& $-1.31$ \;($0.19$) 
& $p=0.34$ 
& $p=0.30$ 
& $-0.12\;|\;-0.07$ 
& $0.29$ \\
\bottomrule
\end{tabular}
}
\caption{Per-item topicality comparison (LLM-based topicality score). All tests are non-significant after Holm; effect sizes are small.}
\label{tab:app-significance-llm}
\end{table}

\medskip\noindent\textbf{Integrated interpretation and consistency with main text.}
Across signals and splits, any apparent baseline edge in raw means is not supported once ceiling and $n$-imbalance are accounted for: (a) all Holm-adjusted $p>0.05$ (the smallest adjusted $p$ observed is $\approx 0.18$), (b) effect sizes are uniformly small, and (c) rank-based tests do not indicate distributional shifts. This aligns with the main-text statement that “after Holm, all $p>0.05$ (min $\approx 0.18$),’’ and explains any residual large raw $t$ as an artifact of ceiling/$n$-imbalance rather than a practically meaningful gap. We therefore treat topicality as \emph{matched} and focus the discussion on difficulty control, diversity, and validity.

\medskip\noindent\textbf{Reproducibility.}
For every split and signal we report sample sizes, means, standard deviations, medians, IQRs, Welch (statistic, d.f., $p$), MWU ($U$, tie-corrected $p$), BM (statistic, $p$), Hedges’ $g$, Cliff’s $\delta$, and Holm-adjusted $p$. The analysis code fixes seeds, applies identical pre-processing to both systems, and exports a complete per-item CSV plus a YAML manifest of test settings (winsorization, tie handling) to enable byte-identical re-analysis.

\medskip
\noindent\emph{Note (ceiling/$n$-imbalance).} Due to the combination of sample-size imbalance (e.g., $\sim$100 baseline vs.\ $\sim$1{,}000 in \textsc{KNIGHT}) and near-zero baseline variance (ceiling effects), Welch’s $t$ can inflate despite negligible practical gaps; we therefore ground conclusions in effect sizes, rank-based tests, and Holm-adjusted decisions.

\subsection{Visualizing Entropy Distributions}

\begin{figure}[h]
    \centering
    \includegraphics[width=0.95\linewidth]{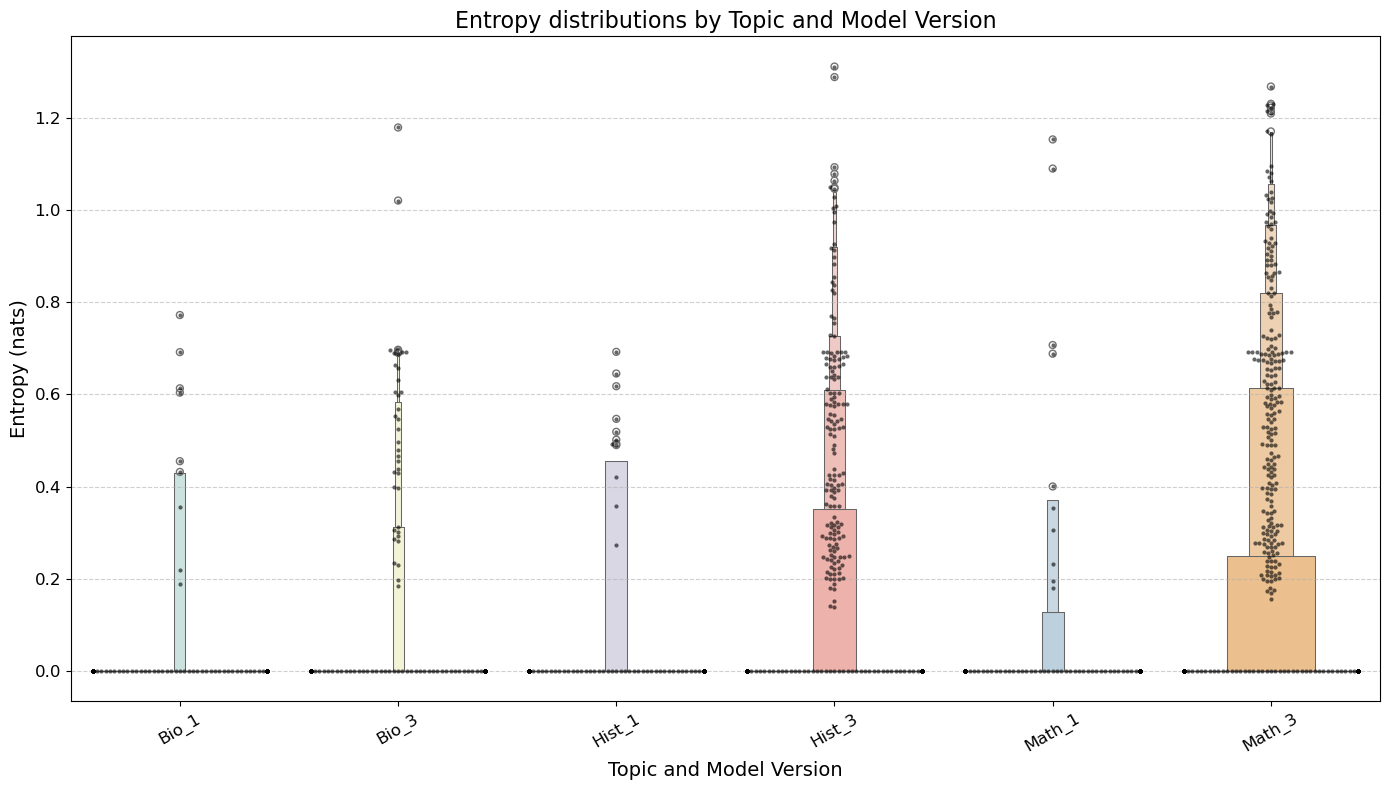}
    \caption{Entropy distributions by topic and difficulty level visualized using boxen plots with swarm overlays. Difficulty level 3 datasets consistently show higher entropy and wider distributions, reflecting greater model uncertainty.}
    \label{fig:entropy_boxen}
\end{figure}

We utilize a boxen plot combined with a swarm plot overlay (Figure~\ref{fig:entropy_boxen}) to visualize entropy distributions across topics and difficulty levels. This visualization method effectively displays not only central tendencies and spread but also highlights data density and outliers in a granular manner.

The plot reveals several important insights:

\begin{itemize}
    \item \textbf{Higher Median and Spread at Difficulty Level 3:} Across all domains, the median entropy and interquartile ranges for difficulty level 3 datasets are notably larger than those for level 1, consistent with increased uncertainty in model predictions on harder questions.
    \item \textbf{Domain Variation in Entropy:} Biology datasets exhibit relatively lower entropy values overall, which aligns with the statistical tests showing insignificant entropy differences between difficulty levels in this domain. History and Math domains display substantially higher entropy at difficulty level 3, indicating a clear gradation of complexity.
    \item \textbf{Presence of Outliers and Data Density:} The swarm plot overlay reveals the distribution density and presence of multiple outliers with unusually high entropy values, particularly in the higher difficulty datasets. This suggests that a subset of questions poses exceptional uncertainty for the model, possibly due to ambiguous or complex content.
    \item \textbf{Skewness in Distribution:} Especially in the Math-3 dataset, entropy distributions show positive skewness, indicating a heavier tail toward higher uncertainty values, reinforcing the notion of increased challenge in these questions.
\end{itemize}

This visualization therefore provides a nuanced and rich depiction of how difficulty modulates model uncertainty, confirming and extending conclusions drawn from entropy statistics and statistical hypothesis tests.

\section{Qualitative MCQ Examples}\label{app:examples}
In this appendix, we present representative multiple-choice questions generated by our system, illustrating how questions are grounded in the constructed knowledge graph (see Figure \ref{fig:exmple1} for a question generation example). Each example consists of the question’s data and the underlying knowledge route. These examples span diverse domains and are arranged to demonstrate increasing reasoning complexity as the length of the knowledge graph path (number of hops) grows from one to three. These exemplars can clarify on how our system integrate factual triples to formulate coherent questions of varying difficulty, with the correct answer highlighted in bold.

\newtcolorbox{bioL1}{
    colback=bioBack,
    colframe=bioFrame,
    fonttitle=\bfseries,
    title=Biology – Level 1
}
\begin{bioL1}
\small{\textbf{Route:}}\\

\texttt{"medicine \& biology"} $\xrightarrow{\textsf{\small INCLUDES}}$  \texttt{"Pharmacology"}\\

\textbf{Question:}\\

Which field of study is included in the comprehensive
overview of medicine and biology, focusing on the effects of drugs on
biological systems?\\
A. Toxicology \\
B. Microbiology \\
\textbf{C. Pharmacology} \\
D. Immunology

\end{bioL1}

\newtcolorbox{bioL2}{
    colback=bioBack,
    colframe=bioFrame,
    fonttitle=\bfseries,
    title=Biology – Level 2
}
\begin{bioL2}
\small{\textbf{Route:}}\\

\texttt{"medicine \& biology"} $\xrightarrow{\textsf{\small UTILIZES}}$
\texttt{"biological insights"} $\xrightarrow{\textsf{\small INFORMS}}$
\texttt{"medicine"}\\

\textbf{Question:}\\

What comprehensive academic overview provides detailed insights into the definition, scope, domains, and applications of both medicine and biology?\\
A. Brief medical summary \\
B. Textbook on pharmacology \\
\textbf{C. Medicine and Biology Overview} \\
D. Public–health article

\end{bioL2}

\newtcolorbox{bioL3}{
    colback=bioBack,
    colframe=bioFrame,
    fonttitle=\bfseries,
    title=Biology – Level 3
}
\begin{bioL3}
\small{\textbf{Route:}}\\

\texttt{"Biomedicine"} $\xrightarrow{\textsf{\small APPLIES}}$
\texttt{"biological principles"} $\xrightarrow{\textsf{\small ENCOMPASS}}$
\texttt{"genetics..."} $\xrightarrow{\textsf{\small IS\_A\_BRANCH\_OF}}$
\texttt{"biology"}\\

\textbf{Question:}\\

Which branch of medical science integrates biological and physiological principles into clinical practice?\\
A. Pharmacology \\
B. Genetics \\
\textbf{C. Biomedicine} \\
D. Immunology

\end{bioL3}

\newtcolorbox{histL1}{
    colback=histBack,
    colframe=histFrame,
    fonttitle=\bfseries,
    title=History – Level 1
}
\begin{histL1}
\small{\textbf{Route:}}\\

\texttt{"Social History"} $\xrightarrow{\textsf{\small EXAMINES}}$
\texttt{"societal structures"}\\

\textbf{Question:}\\

Which historical subfield focuses on the experiences and perspectives of ordinary people?\\
A. Economic \\
B. Political \\
C. Cultural \\
\textbf{D. Social}

\end{histL1}

\newtcolorbox{histL2}{
    colback=histBack,
    colframe=histFrame,
    fonttitle=\bfseries,
    title=History – Level 2
}
\begin{histL2}
\small{\textbf{Route:}}\\

\texttt{"context"} $\xrightarrow{\textsf{\small IS\_VITAL\_IN}}$
\texttt{"archaeology"} $\xrightarrow{\textsf{\small FOCUSES\_ON}}$
\texttt{"human history"}\\

\textbf{Question:}\\

What term refers to the surrounding circumstances that influence the interpretation of human history in archaeology?\\
A. Background \\
B. Setting \\
\textbf{C. Context} \\
D. Environment

\end{histL2}

\newtcolorbox{histL3}{
    colback=histBack,
    colframe=histFrame,
    fonttitle=\bfseries,
    title=History – Level 3
}
\begin{histL3}
\small{\textbf{Route:}}\\

\texttt{"Ottoman Empire"} $\xrightarrow{\textsf{\small ENTERED}}$
\texttt{"World War I"} $\xrightarrow{\textsf{\small ON\_THE\_SIDE\_OF}}$
\texttt{"Central Powers"} $\xrightarrow{\textsf{\small OPPOSED}}$
\texttt{"Allied Powers"}\\

\textbf{Question:}\\

Which major empire entered World War I on the side of the Central Powers, opposing the Allied Powers?\\
A. Austro-Hungarian Empire\\
B. German Empire\\
\textbf{C. Ottoman Empire}\\
D. Russian Empire

\end{histL3}

\newtcolorbox{mathL1}{
    colback=mathBack,
    colframe=mathFrame,
    fonttitle=\bfseries,
    title=Mathematics – Level 1
}
\begin{mathL1}
\small{\textbf{Route:}}\\

\texttt{"fundamental branch"} $\xrightarrow{\textsf{\small REFERS\_TO}}$
\texttt{"arithmetic"}\\

\textbf{Question:}\\

What fundamental branch of mathematics studies numbers and basic operations?\\
A. Algebra \\
B. Geometry \\
C. Calculus \\
\textbf{D. Arithmetic}

\end{mathL1}

\newtcolorbox{mathL2}{
    colback=mathBack,
    colframe=mathFrame,
    fonttitle=\bfseries,
    title=Mathematics – Level 2
}
\begin{mathL2}
\small{\textbf{Route:}}\\

\texttt{"linear algebra"} $\xrightarrow{\textsf{\small PROVIDES\_TOOLS\_FOR}}$
\texttt{"solving systems of linear equations"} $\xrightarrow{\textsf{\small USED\_IN}}$
\texttt{"optimization"}\\

\textbf{Question:}\\

Which branch of mathematics provides essential tools for solving systems of linear equations, a technique frequently used in optimization problems?\\
A. Calculus\\
B. Discrete Mathematics\\
\textbf{C. Linear Algebra}\\
D. Probability Theory

\end{mathL2}

\newtcolorbox{mathL3}{
    colback=mathBack,
    colframe=mathFrame,
    fonttitle=\bfseries,
    title=Mathematics – Level 3
}
\begin{mathL3}
\small{\textbf{Route:}}\\

\texttt{"eigenvectors"} $\xrightarrow{\textsf{\small ARE\_KEY\_CONCEPTS\_IN}}$
\texttt{"linear algebra"} $\xrightarrow{\textsf{\small IS\_A\_BRANCH\_OF}}$
\texttt{"mathematics"} $\xrightarrow{\textsf{\small INVOLVES}}$
\texttt{"logical reasoning"}\\

\textbf{Question:}\\

Which branch of mathematics involves the study of eigenvectors and is essential for logical reasoning in proofs and theorems?\\
A) Geometry \\
\textbf{B) Linear Algebra} \\
C) Calculus \\
D) Statistics

\end{mathL3}

\section{Graph Update and Curator Mechanism}
\label{app:curator}

In this appendix, we describe KNIGHT’s \emph{Graph Update and Curator Mechanism}, which ensures that the knowledge graph grows in a controlled, non-redundant manner.  We reference both the illustrative traversal in Figure~\ref{fig:exmple1} and the pseudocode of the Curator module in Algorithm~\ref{alg:curator}.

\subsection{Formal Definition of Node Curation}
Let \(G=(V,E)\) be the current directed graph, and let \(t\in V\) be the node under expansion. The relation extraction stage applied to node \(t\) proposes a raw candidate set
\(
R = \{\,t'_1, t'_2, \dots\}
\)
of potential new child nodes. The Curator filters \(R\) to produce a curated subset \(C\subseteq R\) of \emph{unique, relevant} new topics that satisfy the following conditions:
\begin{equation}
\begin{split}
  C = \bigl\{\, t' \in R \mid\; & \forall v \in V,\; \neg\,\mathrm{Equiv}(t', v) \\
                               & \quad \wedge \quad \phi(t') = \textsc{True} \bigr\},
\end{split}
\label{eq:curation}
\end{equation}

where \(\mathrm{Equiv}(t', v)\) holds if \(t'\) is judged semantically equivalent to \(v\), either by exact or normalized string match or by high semantic similarity based on cosine similarity of embeddings.
Curator is also a content filter that validates candidate topics by enforcing:

\begin{enumerate}
    \item \textbf{Object type agreement:} Ensures the candidate's semantic type aligns with Wikidata taxonomy \citep{vrandevcic2014wikidata}.
    \item \textbf{Entailment consistency:} Validates logical consistency between the candidate’s description and the relation via natural language inference (NLI) probes \citep{nie2020adversarial}.
    \item \textbf{Content-policy compliance:} Checks adherence to content guidelines and filters out hallucinated or inappropriate information \citep{rejeleene2024towards}.
\end{enumerate}

Only candidates passing all these checks are retained; others are discarded and flagged for human audit by the \textit{Curator} module. Empirically, this multi-stage filtering prunes approximately 7.6\% of candidate edges across domains (\S\ref{sec:experiments}), substantially improving the quality and answerability of the generated knowledge graph.

\subsection{Knowledge-Graph Construction}
\label{ssec:kg}

Given a seed topic $v_0$, \textsc{KNIGHT} crawls Wikipedia and
populates a property graph
$G=(V,E)$ in Neo4j through a recursive expansion process. For each term (starting with $v_0$), the system generates a comprehensive description. This description generation conditionally utilizes Wikipedia as contextual information if available and deemed relevant by an LLM; otherwise, it relies on the LLM with a structured prompt.
From the generated description, subject-predicate-object triplets are extracted. These triplets identify new potential entities (nodes) and their relationships (edges). The newly identified entities are then treated as new terms, and the description generation and relation extraction process is applied recursively to them. The depth parameter $d$ controls the maximum extent of this recursive expansion from the initial seed topic, effectively defining the scope of the resulting KG:
\begin{equation}
  V_d \;=\;\bigl\{v\in V\;\big|\;
      {\rm dist}\bigl(v_0,v\bigr)\le d\bigr\},
  \label{eq:depth}%
\end{equation}
where ${\rm dist}(\cdot,\cdot)$ denotes the shortest-path distance within the constructed graph.
Depth therefore acts as an \emph{intrinsic hardness knob}: increasing
$d$ introduces longer reasoning chains, echoing multi-hop QA
observations \citep{yang2018hotpotqa}.

\subsection{Graph Update and Curator Mechanism}
\label{app:curator}

In this appendix, we describe KNIGHT’s \emph{Graph Update and Curator Mechanism}, which ensures that the knowledge graph grows in a controlled, non-redundant manner.  We reference both the illustrative traversal in Figure~\ref{fig:exmple1} and the pseudocode of the Curator module in Algorithm~\ref{alg:curator}.

\subsection{Formal Definition of Node Curation}
Let \(G=(V,E)\) be the current directed graph, and let \(t\in V\) be the node under expansion. The relation extraction stage applied to node \(t\) proposes a raw candidate set
\(
R = \{\,t'_1, t'_2, \dots\}
\)
of potential new child nodes. The Curator filters \(R\) to produce a curated subset \(C\subseteq R\) of \emph{unique, relevant} new topics that satisfy the following conditions:
\begin{equation}
\begin{split}
  C = \bigl\{\, t' \in R \mid\; & \forall v \in V,\; \neg\,\mathrm{Equiv}(t', v) \\
                               & \quad \wedge \quad \phi(t') = \textsc{True} \bigr\},
\end{split}
\label{eq:curation}
\end{equation}

where \(\mathrm{Equiv}(t', v)\) holds if \(t'\) is judged semantically equivalent to \(v\), either by exact or normalized string match or by high semantic similarity based on cosine similarity of embeddings. Algorithm \ref{alg:curator} illustrates the process for detecting duplicates and semantic aliases (\(\neg\,\mathrm{Equiv}(t', v)\)). Curator is also a content filter that validates candidate topics by enforcing:

\begin{enumerate}
    \item \textbf{Object type agreement:} Ensures the candidate's semantic type aligns with expected taxonomy.
    \item \textbf{Entailment consistency:} Validates logical consistency between the candidate’s description and the relation via natural language inference (NLI) probes.
    \item \textbf{Content-policy compliance:} Checks adherence to content guidelines and filters out hallucinated or inappropriate information using our designed RAG system.
\end{enumerate}

Only candidates passing all these checks are retained; others are discarded and flagged for human audit by the \textit{Curator} module. If a semantic alias is detected, the candidate node \(t'\) is merged with the existing node \(v\), typically by discarding \(t'\) and re-attributing relations to \(v\). Empirically, this multi-stage filtering prunes approximately 7.6\% of candidate edges across domains (\S\ref{sec:experiments}), substantially improving the quality and answerability of the generated knowledge graph.

\begin{algorithm}[t]
\caption{Curator module: Uniqueness and Alias Filtering} 
\label{alg:curator}
\textbf{Input:} current graph \(G=(V,E)\); topic \(t\); raw candidates \(R\).\\
\textbf{Output:} curated set \(C \subseteq R\).
\begin{algorithmic}[1]
  \State \(C \gets \emptyset\)
  \For{\(\textbf{each } t' \in R\)}
    \State \(s \gets \mathrm{normalize}(t')\)
    \If{\(s \in \{\mathrm{normalize}(v) \mid v \in V\}\)}
        \Statex \quad \textbf{continue} \Comment{duplicate name}
    \EndIf
    \For{\(\textbf{each } v \in V\)} \Comment{semantic alias check}
      \If{\(\cos(\mathbf{e}(t'), \mathbf{e}(v)) \ge \tau\)}
        \State \textbf{merge} \(t'\) \textbf{with} \(v\) \Comment{alias}
        \State \textbf{continue to next} \(t'\)
      \EndIf
    \EndFor
    \State \(C \gets C \cup \{t'\}\) \Comment{unique and valid}
  \EndFor
  \State \Return \(C\)
\end{algorithmic}
\end{algorithm}

Once the curated set \(C \subseteq R\) is determined, the knowledge graph is expanded by adding each new topic \(t' \in C\) as a vertex and linking it to its parent \(t\). Formally, this update is expressed as:
\begin{equation}
\begin{split}
  V(G) &\;\leftarrow\; V(G) \;\cup\; C, \\
  E(G) &\;\leftarrow\; E(G) \;\cup\; \{(t \to t') \mid t' \in C\}.
\end{split}
\label{eq:graph_update}
\end{equation}

This procedure guarantees semantic uniqueness and prevents redundancy by merging semantically equivalent nodes, for example, recognizing that “Second World War” and “World War II” represent the same entity.

Beyond the detection and merging of duplicates and semantic aliases, the Curator further applies critical validation checks to ensure the integrity and relevance of graph expansions. These checks include verifying that the candidate node's semantic type conforms to expected classes based on the Wikidata taxonomy \citep{vrandevcic2014wikidata}, ensuring that the relations and descriptions are logically consistent through natural language inference (NLI) techniques \citep{nie2020adversarial}, and screening for compliance with content policies to exclude hallucinated or inappropriate information using our designed RAG system. \citep{rejeleene2024towards}.

\subsection{Question Generation} 
\label{ssec:gen}

Given a topic–seed node \(v_0\), \textsc{KNIGHT} samples a length-\(d\) path through the knowledge graph:
\begin{equation}
  v_0 \xrightarrow{r_1} v_1 \xrightarrow{r_2} \dotsb \xrightarrow{r_d} v_d,
  \label{eq:chain}
\end{equation}
where edges \((r_i)\) and intermediate nodes \((v_i)\), along with their descriptions, are retrieved from the graph. The facts along this chain are verbalized into declarative sentences and embedded into a few-shot prompt template. An LLM (by default GPT-4o-mini) then generates a multiple-choice question (MCQ) consisting of (i) a question stem, (ii) one correct answer option, and (iii) three plausible distractors often related to knowledge graph concepts \citep{yu2024distractor}.

To increase item diversity without additional graph traversals, \textsc{KNIGHT} produces \emph{both} forward and reverse variants of the path in Equation~\eqref{eq:chain}:
\begin{enumerate}
  \item \textbf{Forward mode} (\(\rightarrow\)): The answer node is \(v_d\), and the question stem is framed from the perspective of the seed node \(v_0\), “moving outward.” For example, for \(d=2\), the question might be: \emph{“Which entity, founded by \(v_0\), later merged with \(v_1\)?”}
  \item \textbf{Reverse mode} (\(\leftarrow\)): The path is traversed backwards, treating \(v_0\) as the correct answer. The stem is phrased around the end node \(v_d\), e.g., \emph{“\(v_d\) traces its origins to which founding entity?”}
\end{enumerate}

Empirically, reverse questions increase model entropy by 15–20\% (Section~\ref{sec:experiments}), serving as an effective difficulty augmentation while maintaining factual grounding. Increasing the path length \(d\) typically leads to progressively harder MCQ templates by requiring the integration of information across more hops in the knowledge graph. All generated MCQs, regardless of direction, are subsequently filtered by rigorous quality criteria.

\begin{figure}[ht]
  \centering
  \includegraphics[scale=0.13]{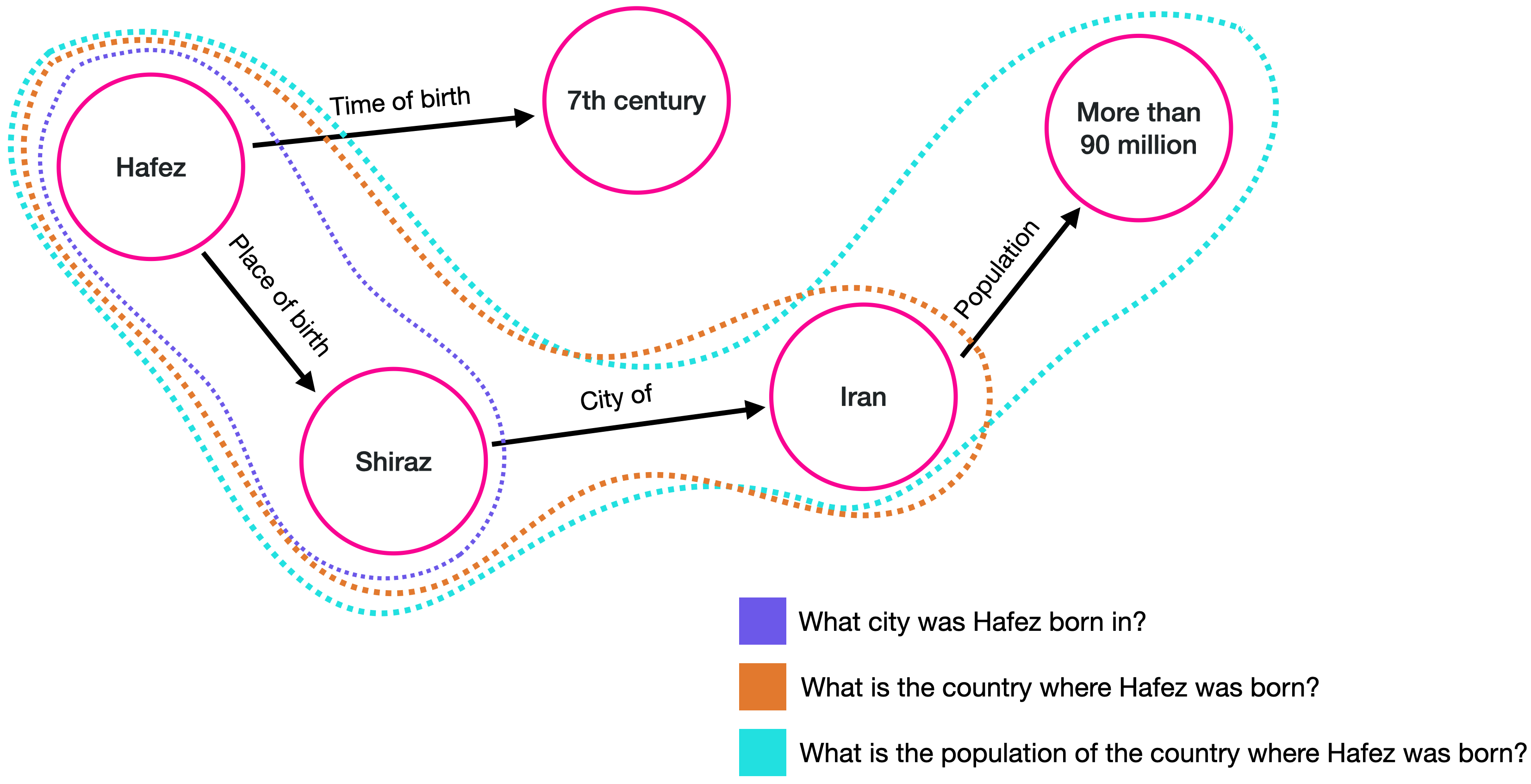}
  \caption{Knowledge‐graph traversal from \textbf{Hafez} to \textbf{Shiraz}, \textbf{7th century}, \textbf{Iran}, and \textbf{>90 million}, generating MCQ templates of increasing difficulty. The purple path denotes a 1‐hop (Level 1) question.}
  \label{fig:exmple1}
\end{figure}

Figure~\ref{fig:exmple1} illustrates a sample knowledge-graph traversal starting from the seed node \textbf{Hafez}. At depth 1 (purple path), the algorithm identifies \textbf{Shiraz} as a connected node. Deeper traversals (not shown) discover nodes such as “7th century,” “Iran,” and “>90 million,” which correspond to progressively harder MCQ templates. The Curator module ensures semantic uniqueness by adding entities like “Shiraz” only once, even if discovered through multiple paths.

\section{Generation Speed and Experimental Hardware Setup}
\label{app:generation-speed}

Our entire QA dataset generation pipeline was executed on Google Colab, utilizing an NVIDIA Tesla T4 GPU with 12 CPU cores. Despite this relatively modest hardware setup compared to large-scale compute clusters, the framework demonstrated efficient runtime performance.

As shown by the empirical measurements in Subsection~\ref{ssec:speed}, KNIGHT enables fast and scalable generation without reliance on expensive or specialized hardware. This efficiency makes it practical for widespread use in research or educational applications, effectively addressing the common time and resource barriers associated with the construction of large-scale QA datasets.

\section{System Parameters and Configuration}
\label{app:parameters}

The behavior and output quality of the \textsc{KNIGHT} framework are influenced by various parameters controlling the underlying models and processes. These parameters are typically configured before running the generation pipeline and allow for tuning performance, creativity, and filtering thresholds. Here, we provide an outline of key parameters and their roles in our system, along with the empirical considerations that led to our default choices.

\subsection{Language Model Inference Parameters}
The Large Language Models used for description synthesis ($\mathcal{L}_{\mathrm{desc}}$), relation extraction ($\mathcal{L}_{\mathrm{rel}}$), MCQ generation ($\mathcal{L}_{\text{q}}$) and validation ($\mathcal{L}_{\text{val}}$) are controlled by the standard generative parameters used primarily. The exact model name is configurable via the environment and key parameters can be configured within the system's source files including:
\begin{itemize}
    \item \textbf{Temperature:} Temperature controls the randomness of the output distribution during the phase of token generation which ranges from 0.0 to 1.0. With the base of empirical testing across different tasks we found that a moderate temperature is beneficial for creative tasks like initial description generation, while a lower temperature is crucial for structured output tasks like triplet extraction where precision is always the first priority. \\
 For initial LLM responses like the description generation task, a moderate temperature of \texttt{0.4} is used to balance creative response generation with factual grounding. Initial tests with higher temperatures led to less coherent text, while lower temperatures reduced desired linguistic variation.\\
For triplet extraction, a lower temperature of \texttt{0.1} is used to encourage more deterministic and focused extraction of structured information. Testing higher values resulted in less reliable triplet formats, failing to consistently adhere to the required structure.
    
    \item \textbf{Max Tokens:} This parameter is set to manage output size and prevent excessive generation costs. In our implementation, the triplet extraction process explicitly sets this to \texttt{2000}. This value was chosen based on testing that showed this limit is generally sufficient to generate all the information needed in the output, while preventing extremely long, potentially irrelevant outputs that could occur without a limit and increase costs or context window issues.
\end{itemize}
The choice of the specific \texttt{OPENAI\_MODEL} (defaulting to \texttt{gpt-4o-mini-2024-05}) was driven by empirical evaluation. Comparative testing against larger models like GPT-4 demonstrated that for our specific tasks and prompt engineering, \texttt{gpt-4o-mini-2024-05} provided a strong balance of accuracy, speed, and significantly lower cost, making it the most practical default for large-scale generation pipelines.

\subsection{Retrieval-Augmented Generation (RAG) Parameters}
The RAG components, primarily utilizing the Wikipedia API, fetch information used for description synthesis. Key parameters influencing this process include:
\begin{itemize}
    \item \textbf{Text Splitting Parameters:} When fetching full Wikipedia page content for processing, a recursive text splitter is used to break the text into manageable pieces. The RAG system is configured with a \texttt{chunk\_size} of \texttt{1000} tokens and \texttt{chunk\_overlap} of \texttt{100} tokens. These values were found to effectively divide the raw text into segments large enough to retain sufficient context and useful data for the LLM while the overlap helps maintain continuity between chunks. 
    Testing smaller sizes sometimes broke up related facts, while larger sizes could exceed context windows or introduce noise without improving retrieval quality.
    \item \textbf{Maximum Returned Chunk Length:} The routine fetching Wikipedia summaries limits the size of the returned text snippet used as context for node description generation. A default limit of \texttt{1000} characters is applied in the code. This value was chosen because empirical observation showed that snippets significantly longer than this rarely added significant value for description synthesis and increased prompt length unnecessarily, potentially diluting the most relevant information for the LLM. Also shorter values deprive the LLM from useful and important context data.
    \item \textbf{Number of Search Results Checked:} When the system needs information about a concept, it searches Wikipedia. This parameter limits the number of top search results from this initial search that the system will look at more closely for relevance before start fetching content from each page which has \texttt{5} as the default value. Checking fewer than 5 often missed relevant pages which results in limiting the scope of the generated graph, while checking significantly more added considerable latency and cost without a proportional increase in the retrieval of highly relevant content.
\end{itemize}
These parameters try to retrieve relevant, concise, and contextually appropriate information to ground the generated descriptions and optimize the process both in quality and efficiency. 

\subsection{Graph Construction and Curator Parameters}
Only one parameter involves in the KG construction and curation process:
\begin{itemize}
    \item \textbf{Maximum Branches:} \allowbreak This parameter controls how many new triplets are extracted from each main idea's (node's) description during the recursive graph expansion with \texttt{2} as the default value. Through empirical testing, we explored limiting branches to 1, 2, 3 or 4. Limiting to 1 often resulted in a shallow, less interconnected graph structure with missing important related concepts. Allowing more than 2 branches significantly increased computational cost and graph size without consistently yielding a proportional increase in the quality or relevance of generated question paths and sometimes introduced noise.
\end{itemize}

\subsection{Validation and Filtering Parameters}
Parameters governing the validation and filtering of generated QA pairs include:
\begin{itemize}
    \item \textbf{Validation Sample Rate:} This parameter allows specifying a sample rate ranging from 0.0 to 1.0 for LLM-based validation of generated QA pairs. 1.0 is the default used for complete coverage in our experiments reported here to provide the best possible quality, but the parameter was added to enable flexibility for situations where speed or cost is more important than verifying each and every pair generated.
\end{itemize}

By tuning these parameters, users can adjust the trade-off between generation speed, dataset size, linguistic style, and the nuances of difficulty calibration for specific domains and use cases. The default values in our experiments represent a balance to produce high-quality, difficulty-calibrated datasets efficiently.

\end{document}